\documentclass[preprint,12pt,authoryear]{elsarticle}

\usepackage{amssymb}
\usepackage{amsthm}

\usepackage{multirow, multicol}
\usepackage{booktabs}
\usepackage{makecell}
\usepackage{pifont}
\usepackage{lscape}
\usepackage{siunitx}

\usepackage[acronym,nomain,nonumberlist]{glossaries}
\newacronym{ann}{ANN}{artificial neural network}
\newacronym{cnn}{CNN}{convolutional neural network}
\newacronym{dnn}{DNN}{deep neural network}
\newacronym{esd}{ESD}{Equivalent Spherical Diameter}
\newacronym{fair}{FAIR}{findable, accessible, interoperable, reusable}
\newacronym{gan}{GAN}{Generative Adversarial Network}
\newacronym{gmm}{GMM}{Gaussian mixture model}
\newacronym{hpc}{HPC}{high-performance computing}
\newacronym{ifcb}{IFCB}{Imaging FlowCytobot}
\newacronym{isiis}{ISIIS}{In-situ ichthyoplankton imaging system}
\newacronym{lbp}{LBP}{Local Binary Pattern}
\newacronym{pwspc}{PWSPC}{Prince William Sound Plankton Camera}
\newacronym{rf}{RF}{random forest}
\newacronym{rdf}{RDF}{random decision forest}
\newacronym{som}{SOM}{self-organizing map}
\newacronym{svm}{SVM}{support vector machine}
\newacronym{pca}{PCA}{principle component analysis}
\newacronym{vit}{ViT}{Vision Transformer}
\newacronym{mlp}{MLP}{multilayer perceptron}

\begin{document}

\begin{frontmatter}



\title{Survey of Automatic Plankton Image Recognition: Challenges, Existing Solutions and Future Perspectives}

\author[lut]{Tuomas Eerola}
\author[lut]{Daniel Batrakhanov}
\author[lut]{Nastaran Vatankhah Barazandeh}
\author[syke]{Kaisa Kraft}
\author[syke]{Lumi Haraguchi}
\author[lut]{Lasse Lensu}
\author[syke]{Sanna Suikkanen}
\author[syke]{Jukka Sepp\"al\"a}
\author[syke]{Timo Tamminen}
\author[lut]{Heikki K\"alvi\"ainen}
\affiliation[lut]{organization={Computer Vision and Pattern Recognition Laboratory, LUT~University},
            addressline={Yliopistonkatu~34},
            city={Lappeenranta},
            postcode={53850},
            country={Finland}}
\affiliation[syke]{organization={Marine Ecology Measurements, Finnish Environment Institute},
            addressline={Agnes~Sjöbergin~Katu~2},
            city={Helsinki},
            postcode={00790},
            country={Finland}}



\begin{abstract}
Planktonic organisms are key components of aquatic ecosystems and respond quickly to changes in the environment, therefore their monitoring is vital to follow and understand the changes in the environment. Yet, monitoring plankton at appropriate scales still remains a challenge, limiting our understanding of functioning of aquatic systems and their response to changes, which reduces the effectiveness of mitigation measures.  Modern plankton imaging instruments can be utilized to sample at high frequencies, producing large amounts of images and enabling novel possibilities to study plankton populations. However, manual analysis of the data is costly, time consuming and expert based, making such approach unsuitable for large-scale application and urging for automatic solutions. The key problem related to the utilization of plankton datasets through image analysis is plankton recognition, i.e., classification of the images. Despite the large amount of research done on automatic plankton recognition, these methods have not been widely adopted for operational use. In this paper, a comprehensive survey on existing solutions for automatic plankton recognition is presented. First, we identify the most notable challenges that make the development of plankton recognition systems difficult and restrict the deployment of these systems for large-scale operational use. Then, we provide a detailed description of solutions for these challenges proposed in plankton recognition literature. Finally, we propose a workflow to identify the specific challenges in new datasets and the recommended approaches to address them. For many of the challenges, applicable solutions exist. This is not the case for all and important challenges remain unsolved: 1) the domain shift between the datasets hindering the development of a general plankton recognition system that would work across different imaging instruments, 2) the difficulty to identify and process the images of previously unseen classes (e.g. plankton species) and non-plankton particles, and 3) the uncertainty in expert annotations that affects the training of the machine learning models for recognition. To build harmonized instrument and location agnostic methods for operational purposes these challenges should be addressed in the future research.
\end{abstract}




\end{frontmatter}


\section{Introduction} \label{introduction}
Plankton, including phytoplankton and zooplankton, is a fundamental component of aquatic ecosystems. They form the basis of the food web and are essential for global biogeochemical cycles~\citep{arrigo2005marine, hays2005climate}. In order to improve management practices of aquatic ecosystems, it is essential to understand functioning of planktonic communities and how they are affected by anthropogenic and climate changes. 

Studying and monitoring plankton is hindered by their microscopic size, fast turnover rates and close interaction with the multiscale hydrodynamics~\citep{benfield2007rapid}. Recent advances in plankton imaging systems have led to their popularization and integration into monitoring programs, collectively accumulating information on plankton systems and simultaneously gathering massive amounts of image data~\citep{benfield2007rapid, cowen2008situ, lombard2019globally, olson2007submersible, picheral2010underwater}. The major constraint to the use of these datasets lies in the expert annotation of plankton images, which is expensive, time-consuming, and error-prone. To fully benefit from the technological development and to properly explore the gathered information, there is a clear need for automated analysis methods.
During recent years, significant research effort has been put into exploring and developing automated methods for performing plankton recognition based on computer vision techniques and machine learning methods \citep[e.g.][]{lumini2019deep, orenstein2017transfer}. 
 
The research on automatic plankton image recognition has matured from early works based on hand-engineered image features combined with traditional classifiers such as \gls{svm}~\citep{cortes1995support} and \gls{rdf}~\citep{ho1995random} \citep[see e.g.][]{tang1998automatic,sosik2007automated} to feature learning-based approaches utilizing deep learning and especially \glspl{cnn}~\citep{lee2016plankton, orenstein2017transfer, lumini2019deep,kloster2020deep}. Various custom methods and modifications to general-purpose techniques have been proposed to address the special characteristics of plankton image data. However, despite the high recognition accuracies reported in the literature, these methods have not been widely adapted to the operational use. The methods that are utilized typically follow rather simple approaches and do not fully exploit the latest advances in computer vision and machine learning. Deploying deep learning based methods for new environments requires large amounts of training data and expert knowledge while publicly available feature engineering based plankton recognition libraries are accessible for non-experts.

Some survey papers on more general microorganism recognition, as well as utilizing machine learning for marine ecology already exist. \citet{zhang2022applications} presented a review of machine learning approaches for microorganism image analysis including history, trends, and applications. The paper covers the segmentation, clustering, and classification of various types of microorganism data. \citet{rani2021machine} described and compared existing microorganism recognition methods. While the challenges are briefly discussed, the discussion remains on a general level and does not go deeply into the solutions. 
\citet{li2019survey} provided a review on microorganism recognition for various different application domains with the focus on traditional feature engineering approaches. 
The survey by \citet{goodwin2022unlocking} covers an even larger scope by addressing the utilization of deep learning methods in marine research. A similar survey was provided by \citet{mittal2022survey}, who presented existing methods on underwater image classification including fish, plankton, coral reefs, seagrass, and submarines. 
\citet{irisson2022machine} provided a plankton recognition review from the application (aquatic research) point-of-view. They present a rather compact survey of the machine learning methods but provide several insights on utilizing machine learning in solving various application-related research questions. \citet{luo2021machine} considered plankton analysis using imaging flow cytometry. In addition to the different imaging technologies, also automatic image analysis methods are reviewed. Those earlier surveys either have considerably wider scope considering various machine learning tasks and organisms, and therefore, not focusing on challenges specific to plankton recognition, or a more narrow scope concentrating on certain technologies for plankton imaging, and thus, lacking a comprehensive review on plankton recognition in general.

In contrast to earlier surveys, we focus on the challenges that researchers commonly face when developing plankton recognition methods and on existing solutions to them. The main goals of this survey are 1) to provide an extensive guide on the available methods to address the challenging characteristics of plankton image data, and 2) to enumerate the challenges that remain unsolved, and which are the most beneficial directions for the future research on the topic. We identify and list the most notable challenges in automatic plankton recognition and provide detailed descriptions of the solutions found in the plankton recognition literature for each challenge. To the best of our knowledge, this paper is the first comprehensive survey focusing exclusively on plankton recognition and the specific challenges related to it.

The rest of the paper is organized as follows. In Section~\ref{Plankton imaging}, the plankton imaging, i.e., imaging instruments and existing image datasets are reviewed. In Section~\ref{Automatic Plankton Recognition}, automatic plankton recognition including feature engineering and \glspl{cnn} are discussed. In Section~\ref{Challenges in plankton recognition}, the most notable challenges in plankton recognition are identified. In Section~\ref{Existing solutions}, the existing solutions for each challenge are described. Finally, the paper concludes with a direction for future research in Section~\ref{Future directions}.

\section{Plankton imaging} \label{Plankton imaging}
\subsection{Imaging instruments}
A fundamental understanding of how plankton species composition is regulated requires frequent and sustained observations. As plankton communities are diverse and dynamic, monitoring plankton is challenging. Different types of plankton imaging and analysis systems have been developed to identify and enumerate living (plankton) and non-living particles in natural waters \citep{benfield2007rapid}. Instruments designed for monitoring plankton communities are briefly discussed next (see review by \citet{lombard2019globally} for more detailed information). The specifications of the imaging instruments are summarized in Table~\ref{table:devices}.

Imaging flow cytometry (IFC) combines 
fluidics, optical characterization and the imaging of cells/colonies. The Imaging FlowCytobot (IFCB) \citep{olson2007submersible} and the CytoSense/Cytobuoy~\citep{dubelaar1999design}, as well as simpler flow systems such as the FlowCam~\citep{sieracki1998imaging} and the ZooCAM~\citep{colas2018zoocam} are among the imaging devices most frequently used within aquatic research. The IFCB is a fully automated, submersible instrument with built-in design features that routinely operate during deployments 
imaging each particle triggering the camera. The CytoSense, available either as a bench top or submersible versions, records forward scatter (FSC), side scatter (SSC) and multiple fluorescence signals of each particle, additionally it can image a subset of the analysed particles. Unlike the IFCB and CytoSense, the FlowCam does not have sheath fluid and it is not an automated in situ instrument. Particle detection in IFCB and CytoSense is triggered by one of the optical sensors (scatter or fluorescence), while FlowCam captures images of a field of view at regular intervals where particles can be identified (autotrigger mode). If the FlowCam is equipped with a laser, particle imaging can be triggered by fluorescence properties, such as the presence of chlorophyll-a. 
The imaging resolution of the IFCB and CytoSense is targeted for a size range of approximately from larger nanoplankton to smaller mesoplankton. The targeted size range for the FlowCam vary according to the combination of flowcell and objective used and instrument versions for imaging of smaller and larger objects and organisms, FlowCam-Nano and FlowCam-Macro, respectively are currently available and image capture is based on autotrigger. The ZooCAM uses an imaging principle similar to that of FlowCam autotrigger.

For obtaining quantitative information from plankton larger than 100 $\mu$m, larger volumes of water are needed to be examined than is possible with IFC~\citep{lombard2019globally}. For imaging of larger particles different types of instruments have been developed utilizing slightly distinct techniques. There are many commercially available instruments such as the In-situ Ichthyoplankton Imaging System (ISIIS)~\citep{cowen2008situ}, Continuous Plankton Imaging and Classification Sensor (CPICS)~\citep{grossmann2015continuous}, ZooScan~\citep{gorsky2010digital},  Video Plankton Recorder (VPR)~\citep{davis2005three}, Underwater Vision Profiler (UVP)~\citep{picheral2010underwater}, and Lightframe
On-sight Keyspecies Investigation (LOKI)~\citep{schulz2010imaging} which are mostly \textit{in situ} imaging systems and their operational principles as well as capabilities are reviewed by \citet{lombard2019globally}. Some instruments have been developed through research purposes but are not commercially available such as the ZooCAM and Prince William Sound Plankton Camera (PWSPC) \citep{campbell2020prince}.

Some of the more recent imaging instruments include the SPC (Scripps Plankton Camera) system~\citep{orenstein2020scripps}, a submersible Digital Holographic Camera (DHC) instrument for temporal and spatial plankton measurements~\citep{dyomin2020monitoring, Dyomin:19}, and its modification, the miniDHC~\citep{dyomin2021underwater, Dyomin:19}. Also HOLOCAM~\citep{nayak2018evidence}, HoloSea~\citep{walcutt2020assessment, macneil2021plankton}, and LISST-Holo are utilized for underwater microscopy using digital holographic imaging (DHI). SPC utilizes an underwater dark-field imaging microscope combined with an onboard computer that allows real-time processing of the images, while the four latter instruments produce 3-D holograms of the imaged volume. The core principal of DHI is in the optical interference phenomenon. A coherent light source, typically a laser, produces the optical interference pattern between undeviated portion of the beam and light diffracted by the object which is recorded on the sensor, and then holograms are reconstructed with pre-/post-processed computer-based algorithms~\citep{WATSON2018106}. The main reasons of emerging DHI microscopy are a wide depth-of-field and field-of-view, i.e., larger sampling volume, and mechanically simpler optical configuration compared to lens-based devices~\citep{walcutt2020assessment, WATSON2018106}.

\begin{table}[ht]
    \begin{minipage}{\textwidth}
    \renewcommand*\footnoterule{} 
    \renewcommand*{\thempfootnote}{\fnsymbol{mpfootnote}} 
    \centering
    \caption{Plankton imaging instruments. For more detailed information about plankton imaging and existing instruments, see~\citet{lombard2019globally}.}
    \label{table:devices}
    \resizebox{\textwidth}{!}{%
    \begin{tabular}{lllllll}
        \toprule    
        \multirow{2}{*}{Imaging instrument} &  \multicolumn{3}{c}{Environment} & \multirow{2}{*}{\makecell[l]{Particle \\ size range}} & \multirow{2}{*}{Image type} \\ \cline{2-4}
         & \textit{In situ} & onboard & laboratory \\
        \midrule
        Imaging FlowCytobot~\citep{sosik2007automated} & \ding{53} & \ding{53} & \ding{53} & 10 - 150\si{\um} & monochrome \\
        \midrule
        CytoSense~\citep{dubelaar1999design} & \ding{53} & \ding{53} & \ding{53} & 1 - 800\si{\um} & monochrome \\
        \midrule
        FlowCam Nano~\citep{sieracki1998imaging} &  & \ding{53} & \ding{53} & 300\si{\nm} - 30\si{\um} & monochrome \\
        \midrule
        \makecell[l]{FlowCam (2x-10x)~\citep{sieracki1998imaging}} &  & \ding{53} & \ding{53} & 3 - 1000\si{\um}~\footnote{depending on the optics} & monochrome/color \\
        \midrule
        FlowCam Macro~\citep{sieracki1998imaging} &  & \ding{53} & \ding{53} & 300\si{\um} - 5\si{\mm} & monochrome/color \\
        \midrule
        ZooScan~\citep{gorsky2010digital} &  &  & \ding{53} & 150\si{\um} - 100\si{\mm} & monochrome \\ 
        \midrule
        LISST-Holo2 & \ding{53} &  &  & 25 - 2500\si{\um} & hologram \\
        \midrule
        UVP-5~\citep{picheral2010underwater} & \ding{53} &  &  & 60\si{\um} - 20\si{\mm} & monochrome \\
        \midrule
        UVP-6LP & \ding{53} &  &  & 60\si{\um} - 20\si{\mm} & monochrome \\
        \midrule
        ISIIS~\citep{cowen2008situ} & \ding{53} &  &  & 60\si{\um} - 130\si{\mm} & monochrome \\
        \midrule
        CPICS~\citep{grossmann2015continuous} & \ding{53} &  &  & 30\si{\um} - 20\si{\mm} & color \\ 
        \midrule
        VPR~\citep{davis2005three} & \ding{53} &  &  & 30\si{\um} - 50\si{\mm} & monochrome video \\ 
        \midrule
        LOKI~\citep{schulz2010imaging} & \ding{53} &  &  & 50\si{\um} - 20\si{\mm} & monochrome \\
        \bottomrule
    \end{tabular}}
    \end{minipage}
\end{table}

\subsection{Publicly available image datasets}
\begin{figure}[h] 
   \centering
   \includegraphics[width=0.65\textwidth]{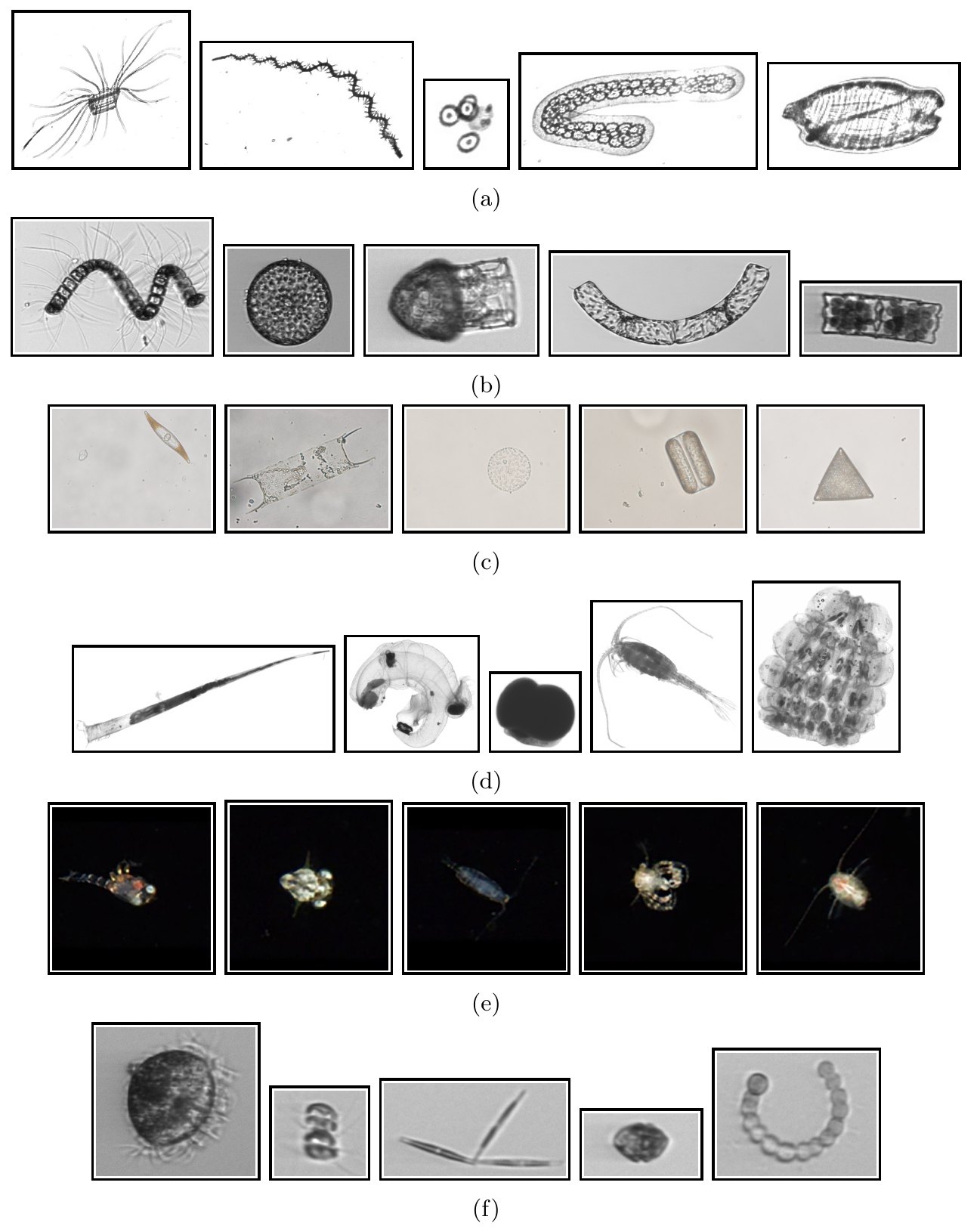}
   \caption{Example images from the publicly available data sets: (a) Kaggle-Plankton~\citep{planktonset}; (b) WHOI-Plankton~\citep{orenstein2015whoi}; (c) PMID2019~\citep{li2019developing}; (d) ZooScan~\citep{ZooScanNet}; (d) DYB-PlanktonNet~\citep{dyb-plankton-Net}; (f) SYKE 2022~\citep{kraft2022syke}.}
   \label{fig:datasets}
 \end{figure}

Publicly available image datasets are crucial on the development of the automatic plankton recognition methods since the most labor intensive part of the process is to create large training and testing datasets. The available datasets are also important for the traceability and comparability of the developed methods. There are several publicly available datasets to be utilized in the research for developing the machine learning methods of plankton recognition. The details of the publicly available and commonly used datasets are summarized in Table~\ref{table:datasets}, and example images from the datasets are shown in Fig.~\ref{fig:datasets}. The most frequently used datasets are ZooScanNet~\citep{ZooScanNet}, Kaggle-Plankton (PlanktonSet-1.0)~\citep{planktonset}, WHOI-Plankton~\citep{orenstein2015whoi,whoi} and their manifold task specific subsets. They all comprise grayscale images collected with a single plankton imaging instrument. UVP5/MC dataset~\citep{UVP5/MC} consists of data collected in the EcoTaxa application~\citep{Picheral2017ecotaxa}. A part of the UVP5/MC dataset has been annotated by an expert and part with an automated tool. More recently collected datasets include PMID2019~\citep{li2019developing}, miniPPlankton~\citep{sun2020few}, DYB-PlanktonNet~\citep{dyb-plankton-Net}, Lake-Zooplankton~\citep{lake_zooplankton}, and the one collected by \citet{plonus2021dataset}. They are acquired with modern imaging instruments and characterized by the presence of color and a higher resolution. SYKE-plankton\_IFCB\_2022 \citep{kraft2022syke} and SYKE-plankton\_IFCB\_Ut\"o\_2021 \citep{kraft2022sykeuto} datasets consist of IFCB images of phytoplankton collected from the Baltic Sea. There are also references to some older commonly used plankton datasets that are not available any more. One example is Automatic Diatom Identification And Classification (ADIAC) database~\citep{du1999diatom}.

\begin{table}[h]
    \begin{minipage}{\textwidth}
    \renewcommand*\footnoterule{} 
    \renewcommand*{\thempfootnote}{\fnsymbol{mpfootnote}} 
    \centering
    \caption{Existing image data sets.}
    \label{table:datasets}
    \resizebox{\textwidth}{!}{%
        \begin{tabular}{llrrlll}
            \toprule
            Dataset & \makecell[l]{Type of plankton } & \makecell[l]{Number of \\ images} & \makecell[l]{Number of \\ classes} & Imaging instrument & \makecell[l]{Data collection \\ region} \\
            \midrule
            \makecell[l]{ADIAC database~\citep{du1999diatom}} & \makecell[l]{Phytoplankton} & 3\,452 & 85 & \makecell[l]{Bright-field microscope} & - \\
            \midrule
            \makecell[l]{Kaggle-Plankton \\ (PlanktonSet-1.0)~\citep{planktonset}} & \makecell[l]{Zooplankton \\ phytoplankton} & 30\,336 & 121 & ISIIS-2 & Straits of Florida, U.S. \\
            \midrule
            WHOI-Plankton~\citep{orenstein2015whoi} & Phytoplankton & over 3.5 M & 103 & IFCB & \makecell[l]{South Beach, Edgartown,\\ Massachusetts, U.S.} \\
            \midrule
            \makecell[l]{UVP5/MC (EcoTaxa)~\citep{UVP5/MC}} & Zooplankton & 1.588 M~\footnote{0.588 M human-annotated and 1 M by MorphoCluster~\citep{schroder2020morphocluster}}  & 65 & UVP5 &  \makecell[l]{Worldwide\\(cruises)} \\
            \midrule
            ZooScanNet~\citep{ZooScanNet} & Zooplankton & 1.433 M & 93 & ZooScan & Villefranche-sur-mer, France \\
            \midrule
            PMID2019~\citep{li2019developing} & Phytoplankton & 10\,819 & 24 & Bright-field microscope & \makecell[l]{Jiaozhou Bay, Qingdao, \\  Shandong, China} \\
            \midrule
            \makecell[l]{miniPPlankton~\citep{sun2020few} \\ (subset of PMID2019)} & Phytoplankton & 1\,400 & 20 & Bright-field microscope & \makecell[l]{Jiaozhou Bay, Qingdao, \\  Shandong, China} \\
            \midrule
            DYB-PlanktonNet~\citep{dyb-plankton-Net} & Zooplankton & 47\,419 & 90  & Dark-field microscope &  \makecell[l]{Daya Bay (DYB), \\ South China Sea, \\ Shenzhen, China.} \\
            \midrule
            Lake-Zooplankton~\citep{lake_zooplankton} & Zooplankton & 17\, 943  & 35  & \makecell[l]{Dual Scripps\\ Plankton Camera} &  \makecell[l]{Lake Greifensee, \\ Switzerland} \\
            \midrule
            Plonus et al. 2021~\citep{plonus2021dataset} & Zooplankton & 218\,000 & 26 & VPR & \makecell[l]{North Sea\\Baltic Sea} \\
            \midrule
            SYKE-plankton\_IFCB\_2022~\citep{kraft2022syke} & Phytoplankton & 63\,000 & 50 & IFCB & Baltic Sea \\
            \midrule
            SYKE-plankton\_IFCB\_Ut\"o\_2021~\citep{kraft2022sykeuto} & Phytoplankton & 57\,000\footnote{Additional 94\,000 images belonging to group unclassified (not possible to assign any of the existing classes).} & 50 & IFCB &  \makecell[l]{Ut\"o, Baltic Sea, Finland} \\            
            \bottomrule
        \end{tabular}}
        \end{minipage}
\end{table}

\section{Automatic Plankton Recognition} \label{Automatic Plankton Recognition}
\subsection{Feature engineering}
A traditional solution for image classification including plankton recognition is to divide the problem into two steps: image feature extraction and classification~\citep{blaschko2005automatic, bueno2017automated, ellen2015quantifying, grosjean2004enumeration, sosik2007automated, zetsche2014imaging,barsanti2021water}. Ideally, image features form a lower-dimensional representation of the image content that contains relevant information for the classification. The main challenge is to design and select good features that are both general and provide good discrimination between the classes. As a result of feature extraction, the obtained feature vectors are used to train a classifier that can then classify unseen images. The most commonly used classifiers for plankton recognition are \acrfull{svm}~\citep{boser1992training, cortes1995support} and \acrfull{rdf}~\citep{ho1995random}. 
\gls{svm} in its most simplistic form is a binary linear classifier that works by mapping the data points in the feature space in such way that the margin between two classes is maximised. It can be extended to multi-class case, for example, by utilizing multiple binary classifiers and to non-linear classification by using a kernel trick.
The \gls{rdf} is a widely used classification method that is based on the observation that combining several classifiers to form an ensemble typically provides better classification performance than any of the individual classifiers. In a typical \gls{rdf}, a large number of decision tree classifiers are constructed and the final classification is obtained by computing the mode of individual classifications. This way, the typical problem of overfitting in the case of decision trees is avoided.

The first work on automatic plankton image classification was presented by~\citet{tang1998automatic}. The image data were produced using a video plankton recorder (VPR)~\citep{davis1992microaggregations} and the proposed method combined texture and shape information of plankton images in a descriptor that is the combination of traditional invariant moment features and Fourier boundary descriptors with gray-scale morphological granulometries. It should be noted that some papers on automatic plankton recognition based on non-image data have been published even earlier. For example, \citet{boddy1994neural} utilized light scatter and fluorescence data obtained by flow cytometry to train an \gls{ann} to classify plankton species. 

Finding good image features is essential for any plankton classification system~\citep{cheng2018review, corgnati2016looking}.
Various feature extraction technologies have been proposed and put into practice for different underwater imaging environments~\citep{sosik2007automated, zetsche2014imaging}. Frequently used plankton features include texture features \citep[e.g.][]{mosleh2012preliminary}, geometric and shape features \citep[e.g.][]{tan2014measuring}, color features \citep[e.g.][]{ellen2015quantifying}, local features \citep[e.g.][]{zheng2017automatic}, and model-based features \citep[e.g.][]{rivas2021fully}. Tables~\ref{table:feature} and~\ref{table:feature2} in Appendix A categorize and summarize various features used for plankton recognition.

The most commonly used image feature type in plankton recognition is shape features \citep[see e.g.][]{sosik2007automated, zetsche2014imaging} that characterize either the contour or binary mask of the object (plankton). In their simplest form geometric features are numerical descriptors of generic geometric aspects such as major and minor axis length, perimeter, equivalent spherical diameter and area of an object computed from binarized image. Another common approach is to utilize image moments to describe the shape. Both Hu moments~\citep{hu1962visual,thiel1995automated,liu2021plankton,zhao2005binary, zhao2010binary} and Zernike moments~\citep{khotanzad1990invariant,blaschko2005automatic} have been proposed for plankton recognition.
Also, various advanced features quantifying the shape of the contour have been proposed for plankton data. These include boundary smoothness \citep[e.g.][]{tang2006binary,liu2020shape}, affine curvature descriptors~\citep{liu2020shape}, Freeman contour code features~\citep{rodenacker2006automatic}, and elliptical Fourier descriptors~\citep{sanchez2019diatom2,beszteri2018quantitative}. Further geometric features applied for plankton recognition include symmetry measures (e.g. Hausdorff distance~\citep{guo2021real,sosik2007automated}) and granulometries~\citep{kingman1975g} utilizing morphological operations~\citep{luo2005active,kramer2005identifying,tang2006binary,wu1998representation}.

Other frequently used type of features in plankton recognition systems are texture features that quantify spatial distribution of intensity or color values in local image regions. While shape features consider only the boundary of plankton, texture features describe the region inside the boundary. The simplest texture features commonly applied in plankton recognition are first-order statistical descriptors that compute simple statistical values directly from the intensity values \citep[see e.g.][]{lisin2006image,zetsche2014imaging,guo2021real}. These are sometimes called color features and include, for example, mean intensity, variance of intensity, as well as, skewness and kurtosis that quantify the shape of the color or intensity histogram. The first order statistics only provide information on how the intensity or color values are distributed in the image. To obtain further spatial information on texture, various second-order statistical descriptors have been proposed. The most common second-order statistical descriptor used in plankton recognition is the co-occurrence matrices~\citep{hu2005automatic,liu2021plankton,shan2020automated,wei2022microalgae}, that describe the statistics of pixel color pairs occurring with certain distance from each other in the image. More advanced texture features proposed for plankton recognition include Local Binary Patterns (LBP)~\citep{ojala2002multiresolution,schulze2013planktovision,chang2016phytoplankton,lisin2006image}, and Gabor descriptors~\citep{idrissa2002texture,sanchez2019diatom,bueno2017automated}.

The third widely utilized group of image features is local features that typically combine the feature detectors and descriptors. Feature detectors search the image for characteristic interest points or regions that contain useful information for the task, i.e. plankton recognition. Local feature descriptors then quantify these regions. General-purpose feature descriptors that have been applied for plankton images include Histogram of Oriented Gradient (HOG)~\citep{dalal2005histograms,bi2015semi,guo2021real}, Scale Invariant Feature Transform (SIFT)~\citep{lowe2004distinctive,tsechpenakis2007image}, Speeded Up Robust Features (SURF)~\citep{bay2006surf,chang2016phytoplankton}, Inner-Distance shape context (IDSC)~\citep{ling2007shape,zheng2017automatic}, and Phase congruency descriptors (PCD)~\citep{kovesi2000phase,sanchez2019diatom,verikas2012phase}.

Feature engineering-based methods for plankton recognition usually combine features from different groups to obtain more representative feature vectors. For example, \citet{zheng2017automatic} used geometric features (e.g. size and shape measurements, such as area, circularity, elongation, convex rate), color features (e.g. sum, mean, standard deviation of color values), texture features (e.g. Gabor descriptors and \gls{lbp}) and local features (e.g. HOG and SIFT). \citet{sosik2007automated} applied simple geometry features, shape and symmetry features, as well as texture features including co-occurrence matrices for phytoplankton recognition.
\citet{wacquet2018combination} extracted 26 features including basic shape features, advanced morphological features, and color features.

Typical plankton recognition systems further apply additional feature selection \citep[see e.g.][]{zheng2017automatic} or dimensional reduction steps to construct compact feature representations. In feature selection, the large set of initial features are ranked based on how representative or informative they are, and the least informative features are discarded. For example, \citet{tang2006binary} proposed normalized multilevel dominant eigenvector estimation (NMDEE) technique to select a best feature set for plankton recognition. In dimensional reduction, \gls{pca} or similar technique is applied to reduce the length of the extracted feature vector while preserving maximum amount of information. For example, \citet{li2013pairwise} and \citet{chang2016phytoplankton} utilized \gls{pca} as a part of the plankton recognition system.

Although feature-engineering-based techniques have been applied with promising results, they require discrete parts, i.e., feature extraction, selection, and training a classifier.
Due to the difficulty of finding general features that provide high classification accuracy over different datasets, feature engineering based plankton recognition methods are often ad-hoc solutions tuned for a single imaging instrument and provide limited accuracy. Moreover, based on previous works~\citep{al2016plankton, khalid2014survey}, it typically requires extensive work to integrate a new class to the existing system. Each new class requires intensive work to find new features that could represent the new class. Depending on the quality of feature design, providing a suitable framework for the accurate, rapid and simplified classification of plankton species is not always possible.

\subsection{Convolutional neural networks}
\begin{figure}[h] 
     \centering
   \includegraphics[width=1\textwidth]{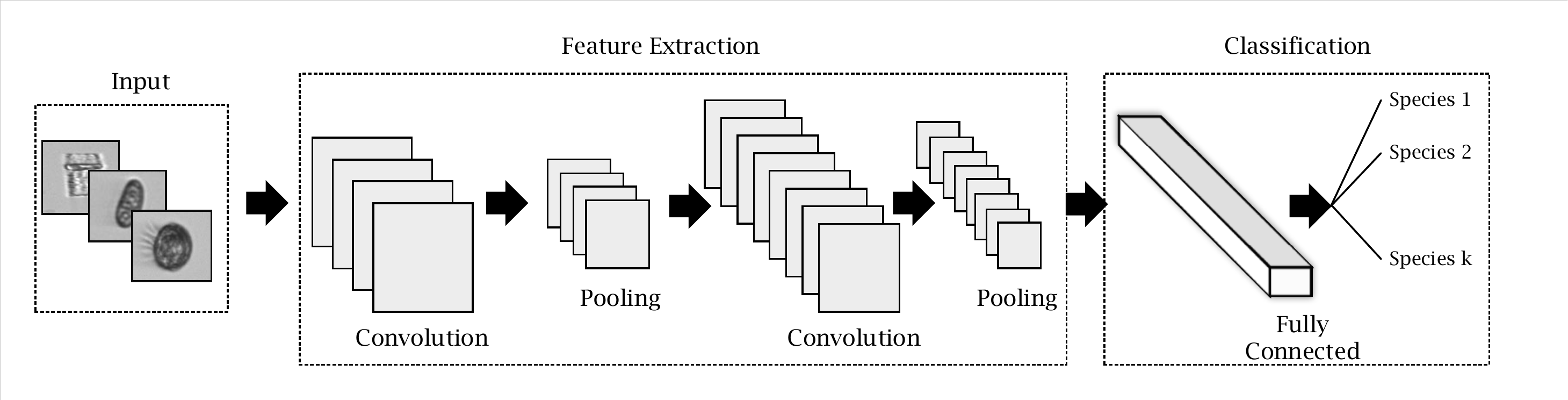}
   \caption{Architecture of a typical convolutional neural network.}
   \label{fig:cnn}
 \end{figure}

\begin{figure}
    \centering
    \includegraphics[width=0.7\textwidth]{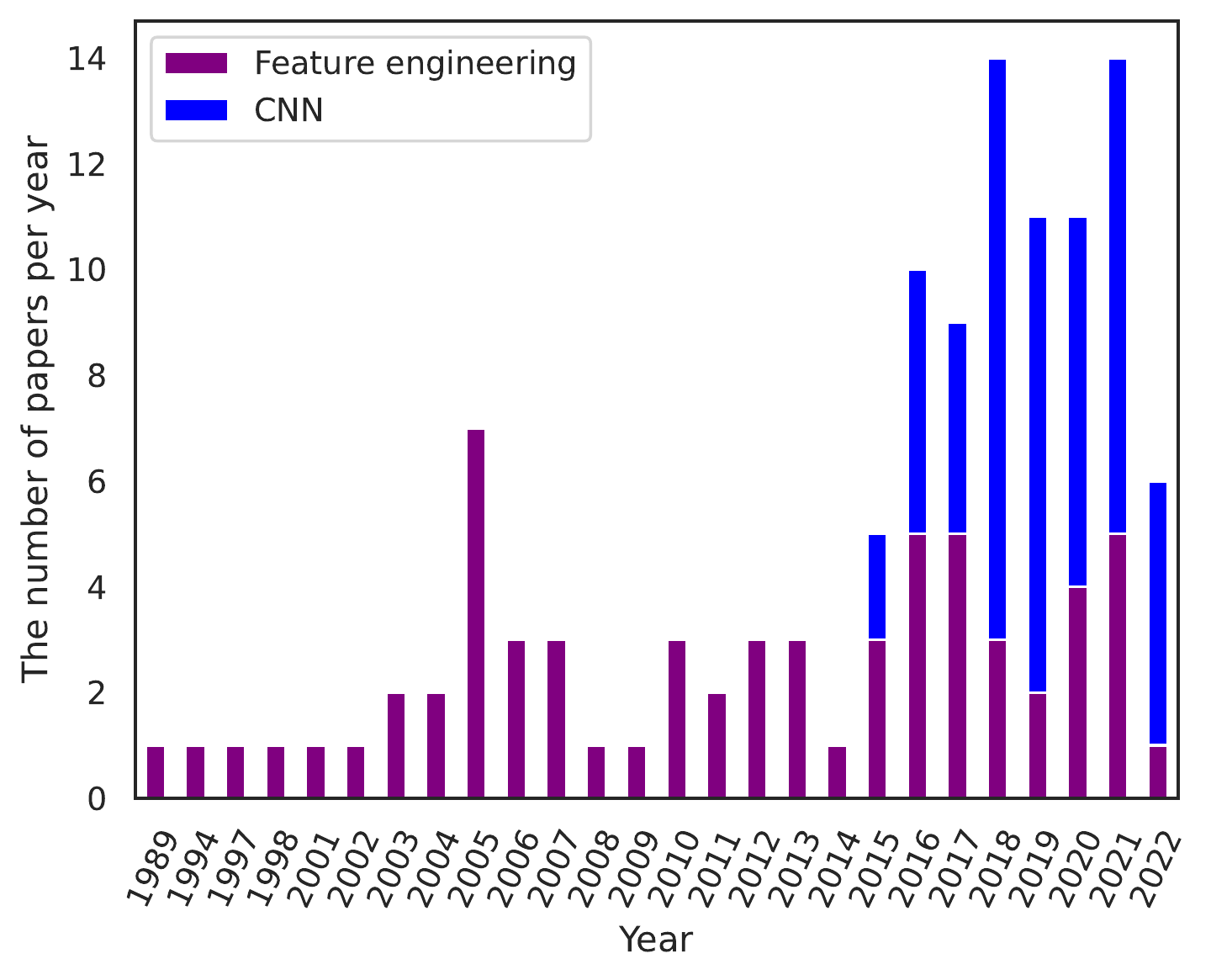}
    \caption{Popularity of feature engineering and feature learning (\glspl{cnn}) based methods on plankton recognition. The plot contains the papers till October 2022.}
    \label{fig:trend_chart}
\end{figure}

%
Recently, \glspl{cnn} have replaced traditional feature engineering techniques in various computer vision applications. The notable difference is that the image features are learnt from the data instead of manually designing them. 
\gls{cnn}~\citep{lecun2015deep} is a type of neural network model for image processing inspired by the animal visual cortex. 
The key component of \glspl{cnn} are the convolutional layers that consist of neurons each processing data only for their receptive field. Due to the shared-weight architecture, these neurons fundamentally perform the convolution operation to the input with a filter defined by the weights of the neurons. This makes it possible to learn the feature extraction filters (weights) through backpropagation. 
A typical \gls{cnn} involves repetitions of several convolution layers and a pooling layer, followed by a set of fully connected layers. The convolution and pooling layers perform feature extraction and the fully connected layers perform the higher-level reasoning and map the extracted features into final output. An example of \gls{cnn} structure is shown in Fig.~\ref{fig:cnn}.
In the recent years \gls{cnn}-based approaches have become dominant in various image analysis tasks providing state-of-the-art performance, for example, in image classification, object localization, and image segmentation tasks~\citep{teuwen2020convolutional}.
Fig.~\ref{fig:trend_chart} illustrates how the popularity of the \glspl{cnn} and feature engineering based approaches on plankton recognition have changed over the years. It can be seen that the introduction of \glspl{cnn} clearly boosted the research in the field. 

The first works considering \gls{cnn}-based classification of plankton images were carried out in 2015. \citet{zheng2015convolutional} carried out preliminary experiments on applying \gls{cnn} for automated plankton recognition. A small \gls{cnn}-model (3-5 layers) was tested on zooplankton data. Similarly, \citet{kuang2015deep} used \gls{cnn} together with data augmentation to solve the recognition task.
\citet{al2015performance} proposed a hybrid solution where \gls{cnn} was used for plankton image feature extraction.

One reason why \glspl{cnn} have become more popular is that they have been shown to outperform the traditional approach utilizing feature engineering multiple times and the architectural components have been studied with care~\citep{gu2018recent}. For example, \citet{zheng2015convolutional} compared a CNN-based plankton image classifier to traditional classifiers such as a multi-layer perceptron (MLP) model utilizing hand-engineered features. The results showed that CNN outperformed the earlier methods.
In various experiments~\citep{orenstein2015whoi,orenstein2017transfer,guo2021real}, \glspl{cnn} have demonstrated higher plankton recognition accuracy than \gls{rdf} combined with hand-selected features.
The preliminary experiments done by \citet{mitra2019automated} suggest that \gls{cnn} can even surpass the human in plankton recognition accuracy.
However, in some special cases, if the computation time is heavily restricted (e.g. embedded systems), feature-engineering based approaches might still be preferable \citep[see e.g.][]{zimmerman2020embedded}.

\subsubsection{\gls{cnn} architectures}
Numerous CNN architectures have been suggested for plankton recognition. 
These include various common \gls{cnn} developed for generic image recognition. For example, \citet{lumini2019deep} compared AlexNet \citep{krizhevsky2012imagenet}, DenseNet~\citep{huang2017densely}, ResNet~\citep{he2016deep}, VGGNet~\citep{simonyan2014very}, GoogleNet~\citep{szegedy2015going}, and SqueezeNet~\citep{iandola2016squeezenet}. DenseNet produced the best classification results with ZooScan, Kaggle-Plankton and WHOI datasets. \citet{liu2018deep} evaluated AlexNet, VGG16~\citep{simonyan2014very},
GoogleNet, PyramidNet \citep{han2017deep} and ResNet. The results suggest that PyramidNet provided improvement on accuracy on a WHOI-Plankton dataset. \citet{sanchez2019diatom} performed a comparison of ResNet, AlexNet, VGGNet, SqueezeNet, DenseNet, and InceptionV3~\citep{szegedy2016rethinking} on a dataset consisting of 1085 diatom images of 14 different classes and DenseNet, ResNet and VGG  provided the highest accuracy.
\citet{kloster2020deep} tested extensively various CNN architectures with fine-tuning. Notably, relatively shallow VGG-16 model outperformed more modern architectures. Table~\ref{table:architecture} in Appendix A gives a summary of different architectures that have been utilized in plankton recognition.

There are also \gls{cnn} architectures developed specifically for plankton recognition. \citet{al2018intelligent} proposed a shallow VGGNet-based architecture for the task.
\citet{dai2016zooplanktonet} proposed a \gls{cnn} architecture called ZooplanktoNet that was characterized by the ability to capture more general and representative features than previous predefined feature extraction algorithms. It was strongly inspired by AlexNet and VGGNet. A comparative experiment with different \gls{cnn} architectures including AlexNet, VGGNet and GoogleNet was carried out and ZooplanktoNet was found to outperform other architectures on zooplankton classification. \citet{li2019tanet} proposed tiny attention network (TANet) consisting of three main parts: a reduction module, self-attention operation, and group convolution. The reduction module was utilized to reduce the information loss caused by pooling operation, self-attention was used to improve the feature learning ability and the group convolution was applied to compress the model size. One of the benefits of the TANet model is its small size which allows real-time classification on mobile devices.
\citet{yan2017more} proposed another light \gls{cnn} architecture for plankton recognition was proposed by utilizing smaller filter size and less fully-connected layers. \citet{luo2021deep} presented a custom architecture MCellNet derived from MobileNetV2~\citep{sandler2018mobilenetv2}. The model was shown to outperform MobileNetV2 on plankton data on both accuracy and computation time. \citet{xu2022accurate} developed a \gls{cnn} for classifying algae based on ResNet and SeNet architectures.

Custom architectures have also been developed for holographic microscopy images as existing image recognition models cannot be directly applied to raw digital holographic microscopy data. A straightforward approach is to first reconstruct images and then utilize any common image recognition architecture \citep[see e.g.][]{qiao2021classification,macneil2021plankton}. This, however, leads to long processing times as the reconstruction stage is computationally heavy. It has been shown that by using a custom architecture \glspl{cnn} can be successfully applied to the raw digital holographic data and the reconstruction step can be avoided~\citep{guo2021automated, zhang2021automatic}.
Also, simulated holograms have been proposed for training and testing simultaneous detection and classification of plankton~\citep{scherrer2021automatic}.

Various works have suggested to use \glspl{cnn} only for the feature extraction and utilize other classifiers, such as \gls{svm} or \gls{rdf} for the final classification step. 
\citet{jindalplankton} suggested to use output of the first fully-connected layer of two \glspl{cnn} (ClassyFireNet and GoogLeNet) as image features and fed it to \gls{rdf} for plankton recognition. Similar approach was evaluated by \citet{orenstein2017transfer} who utilized AlexNet to extract features for an \gls{rdf}-based classifier. \citet{sanchez2019diatom} compared both approaches: fine-tuned \gls{cnn} for classification, and \gls{cnn} for feature extraction. Based on the experiments with various \gls{cnn} architectures fine-tuned \gls{cnn} outperformed the approach where \gls{cnn} was used as feature extractor.

Other commonly used approach is to combine multiple \glspl{cnn} into ensemble to improve the accuracy. This so called ensemble learning is based on the assumption that limited performance of an individual recognition model can be compensated by utilizing additional models more capable of classifying different sets of classes.  
\citet{kuang2015deep} proposed various approaches for model ensemble. These include averaging softmax probabilities and applying principal component analysis for concatenated \gls{cnn} features before softmax classifier. \citet{lumini2019deep} and \citet{lumini2019deepcoral} proposed an ensemble of classifiers by score fusion. Various classifier combinations containing different \gls{cnn} models were evaluated for both plankton and coral classification.
\citet{henrichs2021application} proposed an ensemble of 6 \glspl{cnn} and showed it to outperform an \gls{rdf}-based classifier.
\citet{kyathanahally2021deep} compared various \glspl{cnn} architectures in  ensemble with \gls{mlp} on zooplankton recognition using a mix of feature descriptors and \glspl{cnn} features.  While ensemble learning has shown slightly improved recognition accuracy, it also increases the computation time and complicates the training process.

\subsubsection{Hybrid methods}
Multiple methods that aim to combine the feature engineering approach with \glspl{cnn} have been proposed. One approach is to utilize a separate classifier (e.g. \gls{rdf}) as above. This way \gls{cnn} features can be simply supplemented with selected hand-engineered features before classification \citep[see e.g.][]{orenstein2017transfer,kecceli2017classification}. 
Similarly, ensembles of classifiers can be utilized to combine handcrafted feature based classification and \glspl{cnn}. For example, in the method proposed by \citet{lumini2019deep,lumini2019deepcoral} individual classifiers utilized in the ensembles included various \glspl{cnn} applied to both original images and preprocessed (filtered) images.
The preprocessing techniques included various filters commonly used to compute local features, such as gradient, LBP and wavelets.
\citet{rivas2021fully} combined color and texture features with deep \gls{cnn} features. Both \gls{rdf} and \gls{svm} were tested for classification. 
\citet{dai2016hybrid} proposed a multi-stream \gls{cnn} for plankton classification. In addition to the original image, global feature image representing the shape and local feature image representing the edge information were used as input. All three images were processed with a \gls{cnn} through separate streams. Similar approach was proposed in paper by \citet{cui2018texture}, where the original image, shape image, and texture image were processed in streams before feature concatenation. Concatenated feature maps were processed with one more convolutional and pooling layer, a set of fully connected layers and softmax layer.
A related approach was proposed by \citet{ellen2019improving} who utilized non-image information (metadata) in the \gls{cnn}-based plankton classification. Various architectures to fuse Metadata with \gls{cnn}-based image features were proposed consisting of a set of convolutional and pooling layers for the image and fully-connected layers for the metadata before feature concatenation and common fully-connected layers for the classification.

Also various other modifications to baseline \gls{cnn} classifiers exist. \citet{kosov2018environmental} proposed Conditional Random Field model to utilize spatial relations among pixel-based \gls{cnn} classification results and global features for microorganism detection and recognition. 
\citet{liu2018teaching} proposed to include squeeze-and-excitation block~\citep{hu2018squeeze} to deep pyramidal residual network to increase the plankton recognition accuracy. 
\citet{luo2018automated} took into account the fact that typical plankton images contain a large amount of background pixels without useful information and applied spatially sparse convolutional neural networks originally developed for handwriting recognition~\citep{graham2014spatially}.
\citet{cheng2020method} proposed to combine two \glspl{cnn}, one applied to normal Cartesian coordinate image and one to the same image transformed into Polar representation. This way rotational invariance was obtained in addition to the translation invariance of the baseline \gls{cnn}. 

\subsubsection{Transformers}
In addition to \glspl{cnn}, also other feature learning approaches have been proposed for plankton recognition. One of the most promising approach is \glspl{vit}~\citep{dosovitskiy2021image}, that works by dividing the image into patches resulting in a sequence of vectors (tokens) that are fed to the model. The architecture allows the model to measure relationships between pairs of image patches making it possible to learn to identify the most informative regions in an image via self-attention.
\citet{kyathanahally2022ensembles} applied ensembles of Data-efficient image Transformers (DeiTs) for various ecological image datasets including four publicly available plankton datasets and provided state-of-the-art performance.

\subsubsection{Plankton detection}
Depending on the imaging instrument, there is sometimes a need to first detect the plankton particles in the images~\citep{moniruzzaman2017deep}. Modern \gls{cnn}-based object detection methods such as R-CNN~\citep{girshick2014rich}, YOLO~\citep{redmon2016you}, and their modifications perform the detection and recognition simultaneously, providing end-to-end methods for plankton recognition. For example, \citet{pedraza2018lights} applied R-CNN to detect and classify diatoms in microscopy images, and \citet{soh2018multiple} used YOLO to detect and recognize plankton. \citet{wang2022towards} compared multiple \gls{cnn}-based object detection methods including  Faster R-CNN~\citep{ren2017faster}, SSD~\citep{liu2016ssd}, YOLOv3~\citep{redmon2018yolov3} and YOLOX~\citep{ge2021yolox} on imaging flow cytometer data. YOLOX achieved the best accuracy. \citet{li2021plankton,li2021toward} proposed an improved YOLOv3-based model for plankton detection. The proposed model contains two YOLOv3 networks fused with DenseNet architecture. \citet{kosov2018environmental} applied \gls{cnn}-based images, features and conditional random fields for plankton localization and segmentation.

\subsubsection{Comparison}
Many papers utilize in-house datasets and most publicly available datasets do not provide standardized evaluation protocol meaning that different papers utilize different train-test splits and performance metrics. 
This makes comparison of the performance of different solutions challenging before the principles of making the science \gls{fair} are fully adopted~\citep{schoening2022making}. Table~\ref{table:accuracies} summarizes some published results obtained on publicly available datasets. However, the provided accuracies are not directly comparable due to the reasons mentioned above.
One notable comparison of plankton recognition methods is The National Data Science Bowl~\citep{datasciencebowl} from 2015. The winning team used an ensemble of over 40 convolutional neural networks. 

\begin{table}[t]
    \centering
    \caption{Example accuracies on publicly available datasets from various sources. Numbers should be considered only indicative due to non-standardized evaluation protocols.}
    \begin{small}
    \begin{tabular}{lrrrl}
        \hline
        Dataset & \makecell[l]{Number of\\images} & \makecell[l]{Number of\\species} & Publication: Accuracy \\
        \hline
        WHOI & 3.5 M  & 103 &  \citet{dai2016hybrid}: 96.3\%
            \\& & & \citet{liu2018deep}: 86.3\%
            \\& & & \citet{hoflagellates}: 98.6\%
            \\& & & \citet{venkataramanan2021tackling}: 90.5\%
            \\& & & \citet{guo2021classification}: 72.9\%
            \\& & & \citet{teigen2020leveraging}: 58.8\%
            \\& & & \citet{kyathanahally2021deep}: 96.1\%

    \\
    \hline
        Kaggle- & 30 336  & 121 & \citet{zheng2015convolutional}: 75.726\%  
            \\ Plankton & & & \citet{li2016deep}: 73.1\%
            \\& & & \citet{yan2017more}: 76.4\%
             \\& & & \citet{li2019tanet}: 76.5\%
             \\& & & \citet{geraldes2019situ}: 83\%
             \\& & & \citet{du2020plankton}: 75.8\%
             \\& & & \citet{teigen2020leveraging}: 70\%
             \\& & & \citet{guo2021deep}: 77.45\%
             \\& & & \citet{guo2021classification}: 86.5\%
             \\& & & \citet{kyathanahally2021deep}: 94.7\%

    \\
    \hline
        ZooScanNet & 1.433 M & 93 & \citet{zheng2017automatic}: 88.34\% 
            \\& & & \citet{guo2021classification}: 86.7\%
            \\& & & \citet{kyathanahally2021deep}: 89.8\% 
        \\
        \hline
        ZooLake & 17 943  & 35 & \citet{kyathanahally2021deep}: 98\%                   
    \\
    \end{tabular}
    \end{small}
    \label{table:accuracies}
\end{table}

\section{Challenges in plankton recognition} \label{Challenges in plankton recognition}
\begin{figure}[h] 
     \centering
   \includegraphics[width=\textwidth]{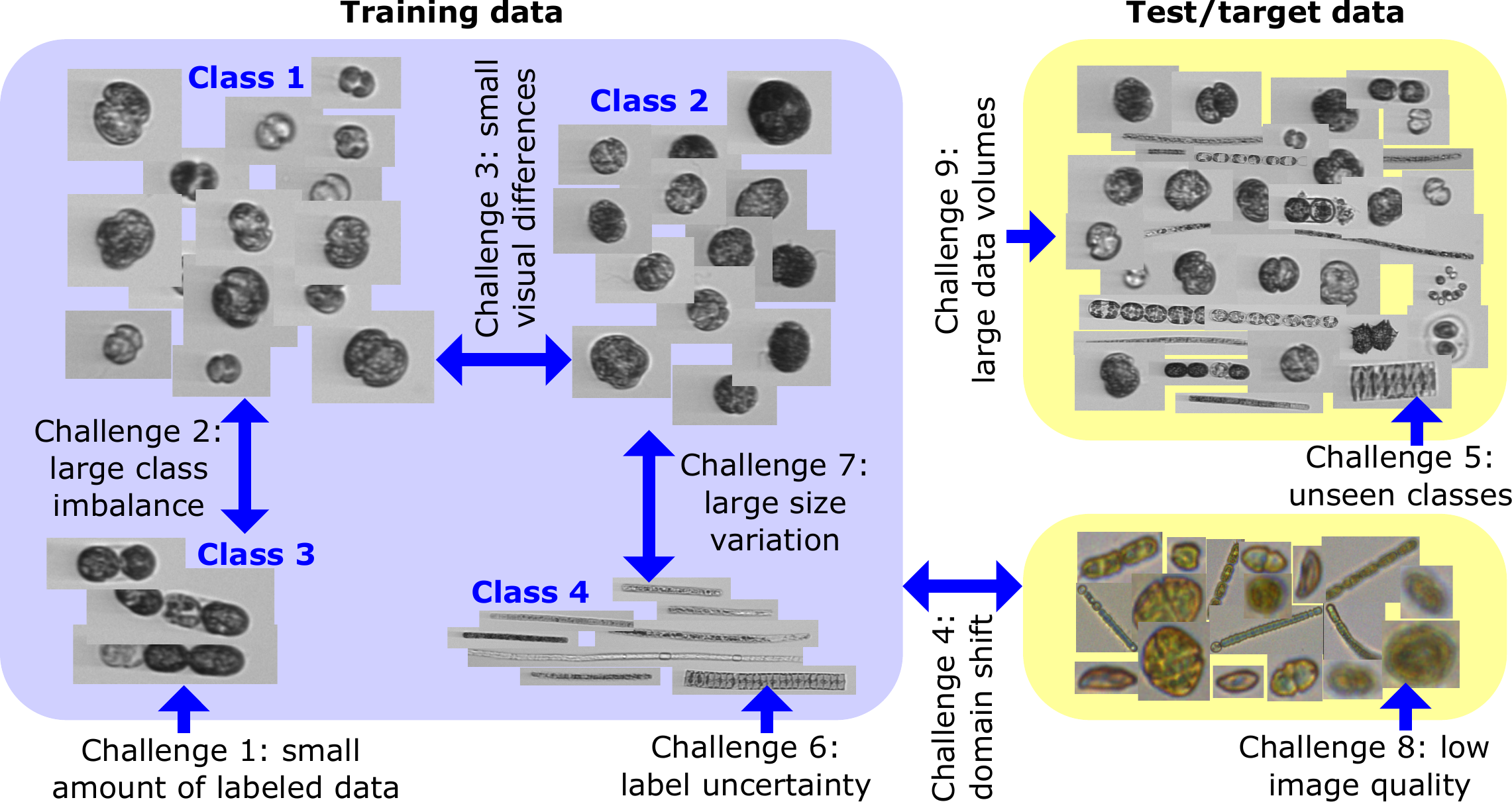}
   \caption{The nine main challenges that complicate the introduction of automatic plankton recognition methods to operational use.}
   \label{fig:challenges}
 \end{figure}

Based on the literature on automatic plankton recognition various challenges can be identified. The most notable challenges are as follows:
\begin{enumerate}

	\item \textit{The amount of labeled data for training is limited.}
	
	This challenge can be divided into two subchallenges: 1) expert knowledge is required for data labeling, and 2) certain plankton species are notably less common producing a small amount of example images.
    Plankton species are inherently difficult to identify, requiring prior expertise. Labeling image data for training and evaluation purposes must be done by experts (e.g. plankton taxonomists) ruling out crowdsourcing tools such as Amazon Mechanical Turk commonly used for labeling large datasets. This makes labeling expensive limiting the amount of labeled data. It also takes years to accumulate enough data to cover rare species.
    Collecting a large training set is essential for deep learning models. Larger amount of training data increases the model's capacity to generalize to new data while training a large model with a small number of examples increases the risk of overfitting, i.e. learning the noise in training data causing the model to perform poorly on unseen images.
    
	\item \textit{There is a large imbalance between classes.}

	Image classification with datasets that suffer from a greatly imbalanced class distribution is a challenging task in the computer vision field. Data of plankton species naturally exhibit an imbalance in their class distribution, with some plankton species occurring naturally more commonly than others. This results in highly biased datasets and makes it difficult to learn to recognize rare species, having a serious impact on the performance of classifiers. 
	Furthermore, with highly unbalanced datasets the overall classification accuracy (e.g. percentage of images that were correctly classified) provides little information about the classes with a small number of samples which may bias the evaluation of the goodness of the classification methods.

	\item \textit{Visual differences between certain classes are small.}

    Certain plankton species, especially those that are taxonomically close to each other, resemble each other visually, which renders the recognition task a fine-grained classification problem. Limitations in the amount of training data make it challenging to ensure that the recognition model learns the subtle differences between the classes reducing the recognition accuracy.

	\item \textit{Imaging instruments vary between datasets.}
    
    If two datasets have been obtained with different imaging instruments producing visually different images (domain shift) the classification model trained on one dataset does not provide sufficient classification accuracy on the other dataset when applied directly. This makes it challenging to develop general-purpose classifiers that could be applied to new datasets limiting the applicability of the existing publicly available large image datasets. There is a need for approaches that allow the adaptation of the trained models to new imaging instruments.

    \item \textit{Training sets do not contain all the classes that can be captured.}

    When deploying a recognition model in operational use, it should be able to handle images from the classes that were not present in the training phase. Different datasets often have different sets of plankton species due to, for example, the geographical distance between the imaging locations or the particle size range of the imaging instruments. Moreover, imaging instruments capture images of unknown particles. Typical CNN-based classification models trained on one dataset tend to classify the images from a previously unseen class to one of the known classes often with high confidence, which not only makes the models incapable to generalize to new datasets and analyze noisy data but makes it difficult to recognize when the model fails. 
    This calls for methods that can identify when the image is from a previously unseen class (species).
    
	\item \textit{There are uncertainties in expert labels.}

    Due to limited imaging resolutions and low image quality, recognizing plankton species is often difficult even for an expert. Manually labeling large amounts of images is tedious work increasing the risk of human errors. Moreover, due to the high costs of labeling work, it is typically not possible to obtain opinions from multiple experts for each image. These reasons cause inaccuracies (uncertainty) in labels to the training data decreasing the classification performance of the trained models. Furthermore, this uncertainty is often highly imbalanced since some of the classes are easier to identify than others.

	\item \textit{Variation in image size and aspect ratio is very large.}

    Most \gls{cnn} architectures require that the input images have fixed dimensions and a typical approach in image classification is to first scale the images into a common size. This is not ideal in plankton recognition due to a very large variation in both the size and aspect ratio of plankton. Scaling images into a common size may cause either small details to be lost in the large images (downscaling) or very large and computationally heavy models (upscaling). Furthermore, the size is an important cue for recognizing the plankton species and this information is lost in scaling. 

    \item \textit{Image quality is often low or has extensive variation.}

    Plankton imaging requires high magnification and the (natural) water might contain other particles, cause unwanted optical distortions, as well as limit the visibility. More importantly, due to the limited depth-of-field, automated imaging instruments often fail to capture particles in focus and the focus may drift away from optimal setting. These reduce the quality of images.
    The low image quality makes both manual labeling (Challenge 6) and automatic classification considerably more challenging. Therefore, there is a need for plankton recognition solutions that are robust to image distortions such as blur and noise.
    
	\item \textit{The amount of image data is massive.}

    Modern plankton imaging instruments produce massive amounts of image data, e.g. FlowCam Macro and ISIIS have the ability to take 10,000 images per minute and 64,000 images per hour respectively. Computationally efficient solutions are needed to perform the analysis in real-time~\citep{macleod2010time, orenstein2015whoi}.

\end{enumerate}
All the nine challenges are visualized in Fig.~\ref{fig:challenges}.

\section{Existing solutions} \label{Existing solutions}
\subsection{Challenge 1: Limited amount of training data}
The two main reasons limiting the amount of training data, the requirement of expert knowledge for the very laborious labeling task and rarity of certain plankton species, require different solutions. 

Active learning has been utilized to minimize the effort of expensive human experts in labeling plankton image data~\citep{luo2005active}. The basic idea behind active learning is to select only the most informative samples for labeling. A classifier is first trained on a small initial training set and the method iteratively seeks to find the most informative samples from an unlabeled dataset. These samples are then labeled by a human expert and the model is re-trained. A simple active learning technique for plankton images called "breaking ties" was proposed by~\citet{luo2005active}. The method utilizes probability approximation for \gls{svm}-based classifier and ranks the unlabeled images based on the differences between the largest and the second largest class probabilities (the smaller the difference the less confident the classifier is). Images with the smallest confidence were labeled by an expert. \citet{drewsjr2013microalgae} studied semi-automatic classification and active learning approaches for microalgae identification. A \gls{gmm} model is estimated from the image feature data and three different sampling strategies are used for the active learning. The experimental results show the benefit of using active learning to improve the performance with few labeled samples. \citet{bochinski2018deep} proposed Cost-Effective Active Learning (CEAL)~\citep{wang2016cost} for plankton recognition. In contrast to traditional active learning where only the manually annotated samples are used in the model training, CEAL utilizes also the unlabeled high-confidence samples for training with class predictions as pseudo labels.
\citet{haug2021combined,haug2021applying,haug2021ciral} proposed Combined Informative and Representative Active Learning technique (CIRAL) to minimize the human involvement in the plankton image labeling process. The main idea behind the method is to find the images with minimal perturbations that are often miss-classified and ignore the images that are far from the decision boundary. The DeepFool algorithm is used to compute small perturbations to the images. The finding of the representative images is formulated as a min–max facility location problem and solved using a greedy algorithm.

\begin{figure}[h] 
     \centering
   \includegraphics[width=0.8\textwidth]{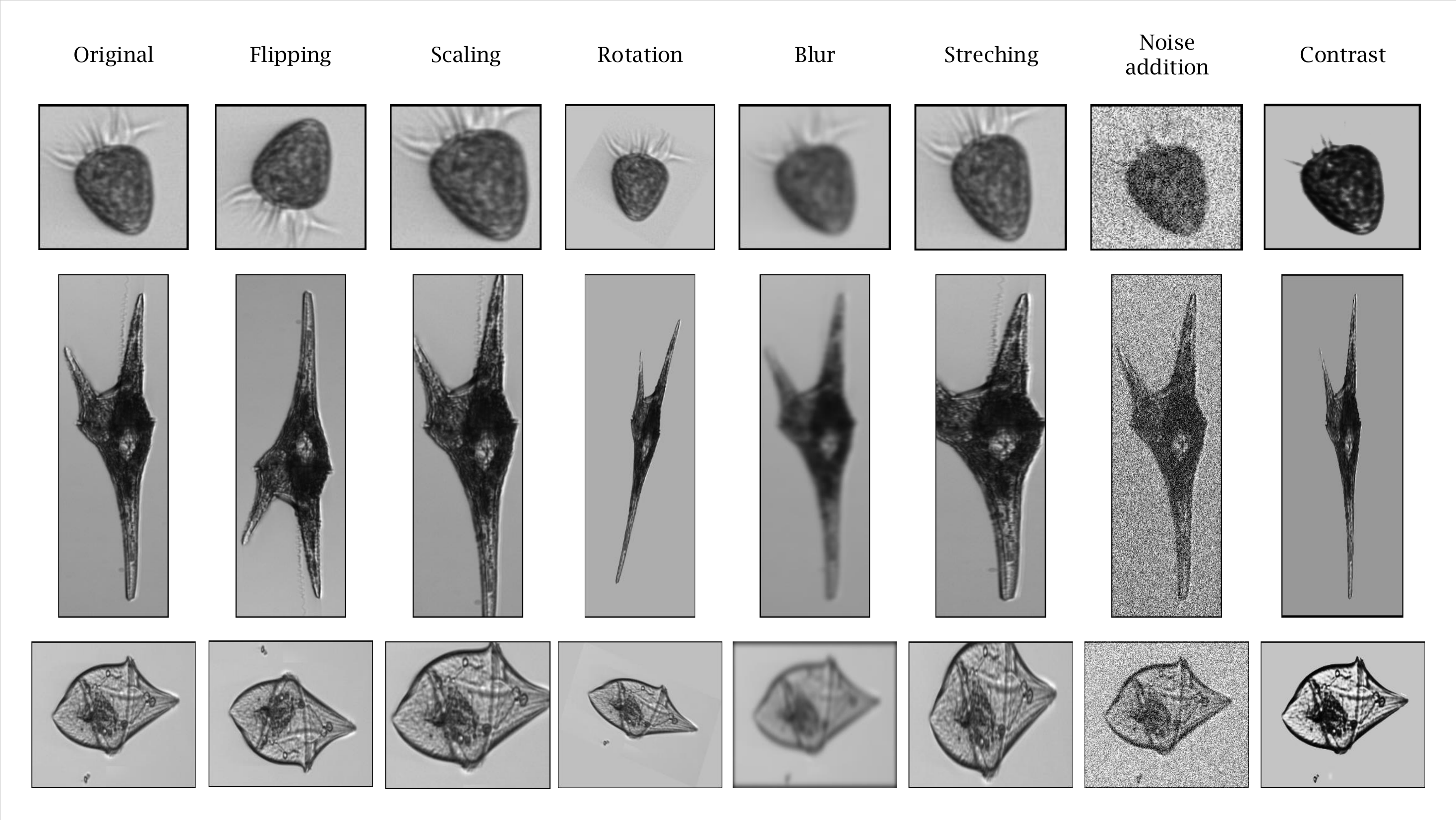}
   \caption{Examples of data augmentation methods.}
   \label{fig:augmentation}
 \end{figure}

While active learning helps to reduce manual work, it is often still a time-consuming process. Typically, there is a need to obtain more training data in a completely automated manner. A traditional approach to increase the amount of training data is to utilize data augmentation. By augmenting the existing labeled image data with various image manipulations, the diversity of the training data, and therefore, the generalizability and accuracy of the trained model can be improved. The most commonly used data augmentation techniques for plankton image recognition include rotation \citep[e.g.][]{cheng2019enhanced, correa2017deep}, shearing \citep[e.g.][]{dai2016zooplanktonet, geraldes2019situ}, flipping \citep[e.g.][]{ellen2019improving, geraldes2019situ}, rescaling \citep[e.g.][]{li2016deep, luo2018automated}, and additional noise \citep[e.g.][]{correa2017deep, geraldes2019situ}. 
Also, blurring~\citep{geraldes2019situ}, contrast normalisation~\citep{geraldes2019situ}, geometric transformations~\citep{orenstein2017transfer,vallez2022diffeomorphic}, as well as adjusting brightness, saturation, contrast, and hue~\citep{dunker2018combining} have been utilized. 
Some works augment images using translation \citep[e.g.][]{dai2016zooplanktonet, li2016deep}. However, it should be noted that \glspl{cnn} are invariant to translation by design, and therefore, this is typically unnecessary when \glspl{cnn} are used for recognition. 
Augmentation has been shown to increase the plankton recognition accuracy even with relatively large training sets \citep[see e.g.][]{song2020classification}.
Examples of augmented images are shown in Fig.~\ref{fig:augmentation}. 

\begin{figure}[h] 
     \centering
   \includegraphics[width=0.8\textwidth]{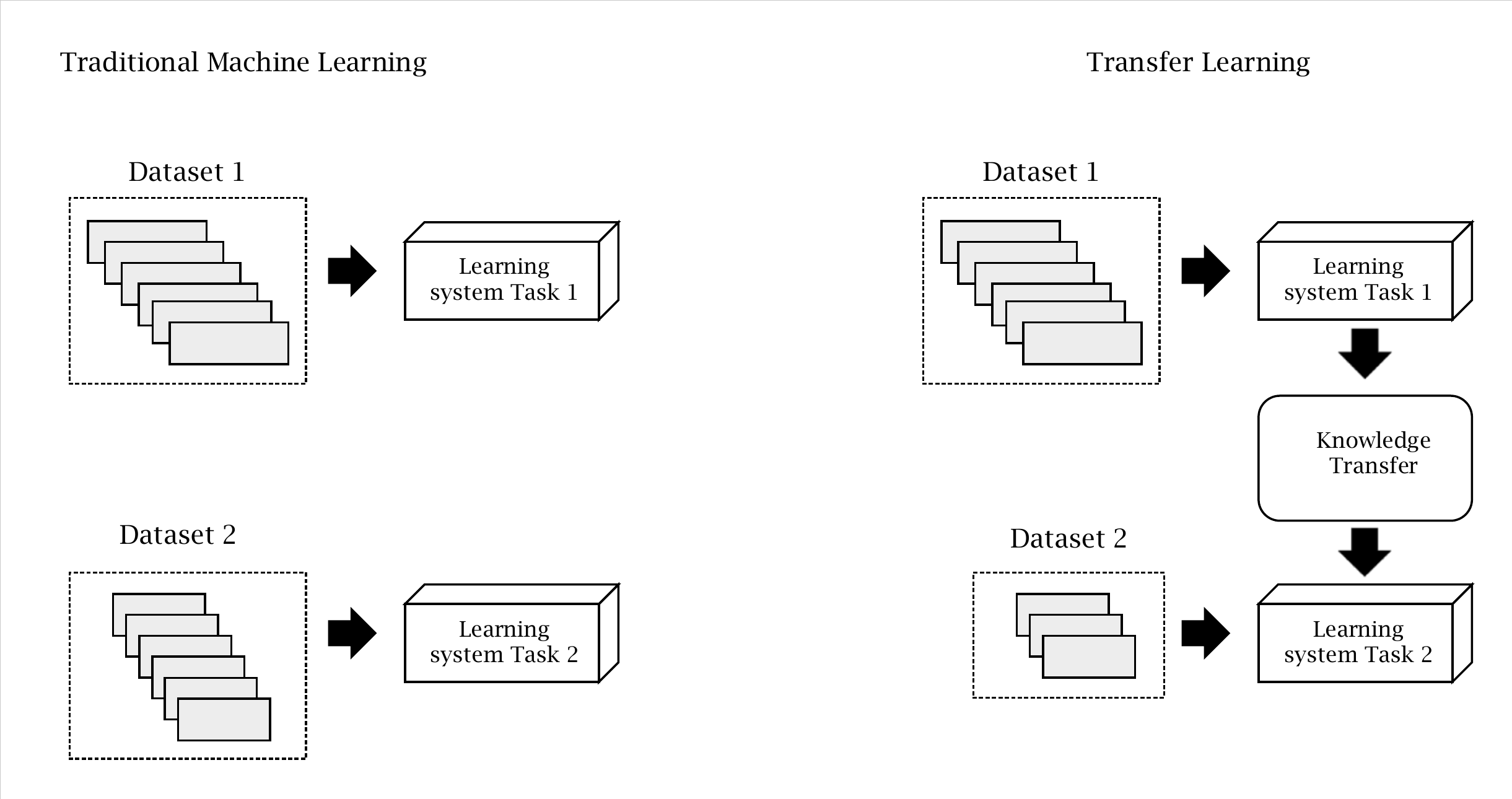}
   \caption{The difference between Traditional machine learning and Transfer learning.}
   \label{fig:transferlearning}
 \end{figure}

Another commonly used approach to address a small amount of training data is transfer learning. Transfer learning is a machine learning method that utilizes knowledge gained from the source domain, where training data are abundant, to the target domain, where training data are scarce~\citep{pan2009survey, shao2014transfer, weiss2016survey} (see Fig.~\ref{fig:transferlearning}).
In the context of plankton recognition, this typically means that the model is first trained using either general image datasets (e.g. ImageNet~\citep{deng2009imagenet}) or a large publicly available plankton dataset and then fine-tuned for the target plankton dataset with typically a limited number of labeled images. Using general image databases as source data is justified by the fact that the learned low level image features are often useful despite the classification problem. In the simplest case transfer learning can be done by simply replacing and training the classification layer and keeping the feature extraction layers unchanged \citep[see e.g.][]{mitra2019automated}. However, it is often beneficial to use the pre-trained network only for initialization and retrain (or fine-tune) the whole network with the target dataset~\citep{lumini2019deepcoral}.

One way to apply transfer learning for plankton images is to use trained \glspl{cnn} only for feature extraction and utilize general classification methods such as \gls{svm} or \gls{rdf} for the recognition \citep[see e.g.][]{rodrigues2018evaluation,rawat2019deep}. However, the results by~\citet{orenstein2017transfer} suggest that better accuracy is obtained by utilizing end-to-end \gls{cnn} with classification layers. \citet{lumini2019deep, lumini2019deepcoral} evaluated various strategies for transfer learning on plankton images. The first strategy was to initialize the model with ImageNet weights and fine-tune the whole model with plankton data. In the second strategy (two rounds tuning), a second pre-training step utilizing out-of-domain plankton image data was added before the fine-tuning. In the third strategy, ensembles of multiple different models were used. Based on the experiments the two rounds tuning did not provide a notable improvement in accuracy.
Similarly, \citet{guo2021deep} explored and compared multiple transfer learning schemes on several biology image datasets from various domains. Various underwater and ecological image datasets are utilized for multistage transfer learning, where ImageNet pretraining is first improved by fine-tuning on an intermediate dataset before, finally, training on the target dataset consisting of plankton images. 
The experimental results show the potential of cross-domain transfer learning even on the out-of-domain data when the number of samples in the target domain is insufficient. 

Large models with more parameters typically require a large amount of data to be trained without overfitting the model. To avoid this and allow the training with a smaller amount of data, shallower \gls{cnn} architectures have been proposed for plankton recognition.
For example, the 18-layer version of ResNet architecture has been shown to achieve a high plankton recognition accuracy on IFCB data \citep{kraft2022towards}. Most custom \gls{cnn} architectures developed especially for plankton recognition including ClassyFireNet~\citep{jindalplankton}, TANet~\citep{li2019tanet}, and ZooplanktoNet~\citep{dai2016zooplanktonet} are relatively shallow with 8, 8 and 11 layers, respectively.
It has been shown that a good classification accuracy could be obtained with a shallow architecture and by using suitable data augmentation methods even with as few as 10 images per class~\citep{kraft2022towards}.

In addition to data manipulation and custom recognition models, also model training approaches have been considered to address the limited data amounts. Learning techniques developed for training the classifier with a minimal amount of samples are called few-shot learning methods. Typically, the idea is to utilize some prior knowledge to allow the generalization to new tasks (in this case classification of new plankton species) containing only a few labeled training examples. Common ways to address few-shot learning is to utilize generation~\citep{hariharan2017low}, embedding or metric learning. The basic idea is to learn such embeddings that the images from the same class are close to each other in the metric space and images from the different classes are far. This allows performing the plankton recognition using distances to the images with known plankton species. Embedding and metric learning have been successfully applied to plankton recognition~\citep{teigen2020leveraging,badreldeen2022open}.

\citet{schroder2018low} employed a low-shot learning technique called weight imprinting~\citep{qi2018low} for plankton recognition with a limited amount of training data. The main idea of weight imprinting is to divide the set of all classes into base classes with enough training data and smaller low-shot classes. During the representation learning phase, a \gls{cnn} is trained to distinguish the base classes with a large amount of training data. In the second phase (low-shot learning), the classifier is then updated with calculated weights to distinguish the smaller low-shot classes. This is done by using appropriately scaled class features of the low-shot classes as their weights, directly allowing the inclusion of classes with only one training image. \citet{guo2021classification} addressed the few-shot learning by supplementing the softmax loss with center loss term~\citep{wen2016discriminative} that forces the samples from the same class close to each other in the deep feature space. The loss function is a weighted sum of the two loss terms and a regularization parameter is used to control the weights. 

In the extreme case, the training data are completely absent and unsupervised learning methods are required. Image clustering is the most commonly used unsupervised technique for plankton image analysis.
\citet{ibrahim2020} carried out preliminary experiments on common clustering algorithms such as k-means with phytoplankton data. Image features for clustering were extracted using pretrained \gls{cnn} models. \citet{coltelli2014water} used various handcrafted image features and self-organizing maps (SOM) for plankton image clustering. \citet{schmarje2021fuzzy} proposed a framework for handling semi-supervised classifications of fuzzy labels due to experts having different opinions. The approach is based on overclustering to identify substructures in the fuzzy labels and a loss function to improve the overclustering. The performance surpassed the one of a state-of-the-art semi-supervised method on plankton data. \citet{salvesen2021unsupervised} studied deep learning for plankton classification without ground truth labels. The improved feature learning was implemented using DeepCluster, a \gls{gan} and a rotation-invariant autoencoder. Despite the potential in unsupervised methods, the gap to supervised learning is still significant.

Hierarchical clustering methods are preferred on plankton data as they have the potential to mimic the taxonomic hierarchy of plankton. \citet{dimitrovski2012hierarchical}, classiﬁcation of diatom images is considered as a hierarchical multi-label classification problem and solved by constructing predictive clustering trees that can simultaneously predict all different levels in the taxonomic hierarchy. These trees are then used as an ensemble forming a \gls{rf} to improve the predictive performance.
Morphocluster~\citep{schroder2020morphocluster} utilizes a semi-automated iterative approach and hierarchical density-based HDBSCAN*~\citep{campello2015hierarchical} for plankton image data analysis. To compute image features for the clustering a \gls{cnn} trained with UVP5/EcoTaxa dataset in a supervised manner was used. The method works iteratively in a semi-automated manner so that clusters are validated by an expert. 
An improved version of Morphocluster was presented by \citet{schroder2022assessing}. Multiple \gls{cnn}-based feature extractors were trained using different labeled datasets to allow the selection of the most suitable feature extractors for the target data. In addition, an unsupervised approach to learn the plankton image features based on the momentum contrast method~\citep{he2020momentum} was proposed. The idea is to use data augmentation to generate two different instances of the same image and use a loss function that forces the model to learn similar feature representations for both instances. Moreover, two custom clustering methods were proposed: 1) shrunken k-Means, and 2) Partially Labeled k-Means. Due to the iterative clustering process of Morphocluster, only part of the images needs to be clustered in each iteration. Shrunken k-Means utilizes distances to cluster centers provided k-means to discard images that are far from the centers. Partially Labeled k-Means utilizes the label information from the earlier iterations to guide the clustering.

Autoencoders have also been proposed for learning plankton image features for clustering without the label information. The basic idea is to utilize encoder-decoder network architecture where the encoder generates an embedding vector from an image and the decoder tries to reconstruct the original image based on the embedding vector. Such a network can be trained without any labels. Ideally, the encoder learns to compress the essential information from the image into an embedding vector that can then be used for clustering. For example, \citet{salvesen2020robust} applied an autoencoder-based approach called Deep Convolutional Embedded Clustering (DCEC) plankton image data. The method employs the \gls{cnn}-based autoencoder architecture by \citet{guo2017deep} and uses k-means to cluster the obtained embeddings.
\citet{alfano2022efficient} proposed a plankton image clustering technique based on variational autoencoders (VAEs). The method utilizes a pre-trained DenseNet without fine-tuning to extract features. Obtained deep image features are then fed to VAE to generate latent space representations. Finally, low-dimensional embeddings are clustered using fuzzy k-means. 

Clustering methods are only able to produce unlabeled clusters of images with a similar appearance. Therefore, further analysis is needed to confirm and label the clusters. \citet{schroder2020morphocluster} addressed this by introducing an interactive tool where the users revise the obtained clusters, manually correct the hierarchy and annotate the final set of clusters. This semiautomatic approach reduces the manual work needed for data labeling as the expert does not need to annotate every image separately. \citet{goulart2021deep} utilized t-distributed stochastic neighbor embedding (t-SNE) to visualize the clusters in two-dimensional space allowing the human expert to quickly see the clusters in the data.
\citet{pastore2020annotation} proposed a full pipeline for environmental monitoring based on plankton image clustering and minimal expert supervision (the expert labels only one image per cluster). \gls{cnn} was used for image feature extraction and various unsupervised clustering algorithms including K-means, fuzzy K-means, and Gaussian mixture model were compared. 

As a summary, the most common approaches to tackle the problem of limited amount of labeled plankton image data are data augmentation and transfer learning. Data augmentation is an essential part of practically all modern plankton recognition pipelines based on deep learning, while transfer learning allows to utilize knowledge from another domain to compensate the lack of training data. In the case of extreme scarcity of labeled training data, further modifications to the model training are needed. Typically this means the adoption of regularization techniques that prevent the model to overfit to the training data. Weight imprinting, metric learning, and central loss have been found useful tools in few-shot plankton recognition. If labeled training data is completely missing, clustering or active learning can be utilized. Clustering allows to analyze plankton image datasets in an unsupervised manner, while active learning makes it possible to minimize the amount of expert labeling effort for building a plankton recognition model for future data.

\subsection{Challenge 2: High class imbalance}
High class imbalance is naturally inherent in many real-world applications and plankton recognition is not an exception. Certain plankton species are considerably more common than others causing the data in typical plankton datasets to be highly imbalanced. This is problematic when it comes to training plankton classification methods. One of the most notable problems connected to the high class imbalance is the catastrophic forgetting where neural network, while learning new information, completely forgets previously learned information. This typically affects the minority classes that are only rarely seen during the training stage causing the network to only learn the necessary image features for the majority classes.

Undersampling is a technique to decrease the level of imbalance by discarding images from the majority classes. In the simplest case, undersampling can be done by randomly selecting a subset of images from majority classes is such way that the resulting training dataset has an equal amount of images in all classes.
For example, \citet{lee2016plankton} reduced the class bias on small-sized plankton classes by randomly sampling images from the classes with more samples than the predefined threshold. \citet{kloster2020deep} utilized a similar undersampling technique.
Also, more intelligent solutions for undersampling have been suggested in plankton literature.
\citet{le2022benchmarking} utilized undersampling by filtering combined with cost-sensitive learning to obtain a more balanced dataset for training.
\citet{ding2018plankton} proposed an EasyEnsemble.D algorithm for plankton recognition on highly imbalanced datasets. The basic idea is to sample multiple subsets from majority classes to fully utilize the large data volumes. Each subset is used to train a separate weak classifier with different weights, and the final classification is performed using the ensemble of the weak classifiers. The problem with undersampling is that it reduces the amount of training data which in the case of plankton recognition is typically already limited. Especially, in the presence of rare species, the undersampling alone leads to an extremely small training set.

Oversampling is another technique to reduce the level of imbalance with duplicating samples from the minority classes. The oversampling is typically done using data augmentation, i.e. instead of using identical duplicates, manipulated versions are created to obtain more training data for minority classes. For example, \citet{bochinski2018deep}, increased the amount of training samples of the smaller classes by mirroring the images horizontally and vertically to counter the imbalance during training.

\citet{xiaoyan2020research} proposed a combination of undersampling and oversampling to address the class imbalance in plankton recognition. This is done by utilizing KA-Ensemble algorithm~\citep{ding2020ka} that combines oversampling of the minority class via kernel-based adaptive synthetic sampling (Kernel-ADASYN) and random undersampling of the majority class.
The experiments showed increased classification accuracy for the minority class.
\citet{liu2021plankton} proposed to combine borderline-SMOTE oversampling with Fuzzy C-means clustering-based undersampling for plankton image data. The Synthetic Minority Oversampling TEchnique (SMOTE)~\citep{chawla2002smote} synthesizes new samples between the minority class and its nearest neighbor in the feature space. Borderline-SMOTE~\citep{han2005borderline} improves the method by concentrating on the samples near the class boundaries in order to oversample more significant samples for the minority classes. Fuzzy C-means clustering is utilized to preserve the clusters found in the original data during undersampling.

Another approach among a variety of resampling methods is cost-sensitive learning~\citep{elkan2001foundations}. The method defines a so-called cost-matrix which specifies a reward or a penalty over the classifications of an algorithm. A core idea behind it is similar to resampling but it does not change the prevalence of the training set directly. However, a performance evaluation for an imbalanced plankton set reported by \citet{correa2016supervised} demonstrates only minor improvements for cost-matrix in comparison to SMOTE and resampling.

Another solution to artificially create more image data for training and to reduce the level of imbalance is to utilize generative models capable of generating realistic images with a certain distribution. \glspl{gan}~\citep{goodfellow2014generative} are deep learning models that can be used to generate photo-realistic artificial images with the same statistics as the data they were trained with. This is done by using two models, a generative model and a discriminative model. The generative model generates candidate images usually from random noise. The discriminative model is an image classifier that is given labeled samples from the real set of images and fake images produced by the generative model. The task of the discriminative model is to distinguish real images from fake ones and the task of the generative model is to fool the discriminative model. These two models are trained simultaneously in such a way that the generative model becomes increasingly better at producing realistic fake images and the discriminative model gets increasingly better at recognizing them. \glspl{gan} have been shown to be able to generate images that are authentic to human observers.

\glspl{gan} have been utilized also for reducing bias caused by the class imbalance in plankton recognition. 
\citet{wang2017cgan} used \gls{gan} to generate new example images of minority classes. Furthermore, a method was proposed where the \gls{cnn}-based plankton recognition model shares the weights with the discriminative model. However, only minor improvement was observed over the baseline recognition models trained on the original data without \gls{gan}-based data augmentation.
\citet{liu2018teaching} proposed a \gls{gan}-based curriculum learning strategy. The proposed method contains two stages. First, the model is trained using the original data and then with more complex data consisting of GAN-generated images.
\citet{li2021plankton} utilized CycleGan~\citep{zhu2017unpaired} for the augmentation of rare taxa, and \citet{khan2022multiclass, ali2022computer} applied DC-GAN~\citep{radford2015unsupervised} to augment an algae image dataset. \citet{vallez2022diffeomorphic} compared data augmentation by combining two diatom images from the same class using morphing and image registration methods performing diffeomorphic transformations to generation of synthetic images by a \gls{gan}. In this study, mixing images using morphing achieved better results.
The fundamental problem of using \glspl{gan} for image augmentation is that the generated images have the same statistics as the images they were trained with. Therefore, if the \gls{gan}s are trained using the same data as the recognition model, and the recognition model is able to learn the data distribution from the original data, the generated samples do not necessarily provide additional value for the training. However, some promising results have been obtained on \gls{gan}-based augmentation of highly imbalanced datasets~\citep{tanaka2019data}.

Similarly to the challenge of a limited amount of training data, transfer learning has also been proposed to overcome the class imbalance problem. 
In a method proposed by \citep{lee2016plankton}, a balanced dataset is first generated using randomized undersampling, the model is pre-trained on the balanced dataset, and finally fine-tuned using the whole unbalanced plankton image dataset. \citet{wang2018transferred} introduced a transfer parallel model approach for plankton recognition. The main idea is to avoid the catastrophic forgetting by training two submodels: 1) a model trained on the whole dataset, and 2) a pre-trained model trained only on small classes. Deep features from both of the models are concatenated before the softmax layer. The latter submodel adds good image features for minor class classification that the network could otherwise fail to learn.

Also, modified model architectures have been proposed to address the class imbalance. These include models with increased generalization ability to minority classes. \citet{liu2018deep} applied Deep Pyramidal Residual Network (PyramidNet)~\citep{han2017deep} to plankton recognition and shown to improve accuracy on a highly imbalanced dataset. The idea behind PyramidNet is to gradually increase the size of the feature map. This combined with the ResNet style to skip connections causes reduced change of overfitting, and therefore, better generalization ability. \citet{kerr2020collaborative} proposed model fusion to address the class imbalance. The results suggest that combining multiple individually trained \glspl{cnn} with a common softmax layer improves the accuracy of rare species, consequently providing better overall accuracy on imbalanced data. 

As a summary, undersampling and oversampling are the simplest and most widely used approaches to address high class imbalance in plankton image data. Oversampling is typically performed using traditional data augmentation, but also generative approaches such as GANs have been proposed to generate completely new plankton images for the minority classes. Moreover, transfer learning, model fusion, and regularization techniques preventing overfitting have been shown to improve plankton recognition accuracy in the case of highly imbalanced training data.

\subsection{Challenge 3: Fine-grained nature of the recognition task}
In order to obtain high recognition accuracy on classes with high inter-class similarity such as taxonomically close plankton species, techniques that focus attention on subtle visual differences are needed. The task of recognizing hard-to-distinguish classes from each other is called fine-grained classification. Plankton recognition in most cases can be considered a fine-grained classification task as the fundamental way to improve the overall accuracy of a recognition model is to make it better at recognizing the challenging cases. Despite this, most of the work on plankton recognition does not tackle the challenge directly but instead focuses on comparing different general model architectures on the task. Related to this viewpoint, it has been also studied whether the recognition should be considered as a flat or hierarchical classification task. \citet{boddy2000identification} considered misclassifications of phytoplankton as a result from the overlap of feature distributions and grouping of similar species within genera or based on groupings indicated in dendrograms was proposed. Similarly, \citet{fernandes2009optimizing} proposed an approach for balancing the trade-off between the classification performance and number of classes. The model automatically suggests merging of classes based on the statistics evaluated after the classification. The results from taxa recognition of macroinvertebrates by \citet{arje2020human} showed that humans performed better when a hierarchical classification approach commonly used by human taxonomic experts was used, but when a flat classification approach was used, the \gls*{cnn} was close to human accuracy. To improve the automatic approaches, a few methods focusing especially on the attention mechanism to address the fine-grained nature of the recognition task have been proposed.

\citet{sun2020few} considered fine-grained classification of plankton by proposing an attention mechanism based on Gradient-weighted Class Activation Maps (Grad-CAM)~\citep{selvaraju2017grad} to force the \gls{cnn} to focus on the most informative regions in the image. Grad-CAM was originally developed for visualizing the \gls{cnn}-based models. It highlights important image regions which correspond to the decision of interest (in this case plankton recognition). \citet{sun2020few} utilized Grad-CAM to detect the regions to focus on, and a feature fusion approach utilizing high-order integration~\citep{cai2017higher} is applied to obtain stronger features for those regions. This approach shares similarities with the self-attention module used in the TANet architecture~\citep{li2019tanet} for plankton recognition. However, the self-attention module puts larger weights on the important regions, i.e. those regions in the feature map with high activation values.

Also other approaches for fine-grained plankton recognition have been proposed. \citet{du2020plankton} applied Matrix Power Normalized CO-Variance (MPN-COV) pooling layer for second-order feature extraction. The aim is to model the complex class boundaries more accurately than in traditional pooling (e.g. softmax). There is some evidence~\citep{li2017second} that suggests that higher-order information can improve recognition accuracy in fine-grained tasks. \citet{venkataramanan2021tackling} proposed an improved pipeline tackling inter-class similarity and intra-class variance. The authors suggested alleviating inter-class variance with a metric learning-based approach utilizing triplet loss and mitigating intra-class variance by X-means clustering technique applied to the extracted features. The idea is to cluster the classes with high inter-class variance into multiple clusters and consider these as separate classes. The authors propose a method to find the optimal amount of clusters that minimize both the intra-class variance and inter-class similarity, and this way improve the accuracy of fine-grained plankton recognition.

In general, only few papers directly tackling the fine-grained nature of the plankton recognition task exist. These are based on attention mechanisms to find the most important regions in the images allowing the recognition model to focus on the subtle differences between the classes, and contrastive or metric learning that allow explicitly learning the image features that separate the pairs of classes. 

\subsection{Challenge 4: Domain shift between datasets}
Different imaging instruments cause domain shift between plankton data-sets. Domain shift in a wider sense refers to a situation where the distribution of the dataset that is used for training differs from the data where the recognition model is applied. \gls{cnn}-based models tend to learn image features that are very specific to the distribution of the training data making them notoriously weak at generalizing beyond the domain they were trained on \citep{gulrajani2020search}. This is why most automatic plankton recognition solutions focus on just one imaging instrument. This, however, limits the wider utilization of the methods. Tuning the classification model trained on one dataset to work on another dataset (correcting domain shift between the datasets) is called domain adaptation \citep{bendavid2010theory} and learning a general model that can be applied to any dataset (domain) is called domain generalization \citep{zhou2022domain}. 

While domain adaptation and generalization have not been widely studied on plankton recognition, there have been works where multiple different plankton image datasets have been utilized to solve the recognition task. 
Transfer learning and fine-tuning have been utilized as approaches against the differences in datasets. \citet{rodrigues2018evaluation} applied transfer learning using \glspl{cnn} to obtain a feature extractor that can be used for new datasets. The Kaggle-Plankton dataset was used to train a \gls{cnn} (source dataset) and an in-house dataset was used as a target dataset to test the suitability of the features.
\citet{orenstein2017transfer} applied a variety of learning schemes to three very different plankton image datasets. The bigger labeled image datasets, IFCB and ISIIS, were used to train \glspl{cnn} both by fine-tuning and from scratch. Then, the classifiers were used to classify within-domain images directly and as feature extractors for out-of-domain data.

\citet{lumini2019deep, lumini2019deepcoral} studied ensembles of different \gls{cnn} models, fine-tuned on several datasets, with the aim of exploiting their diversity in designing an ensemble of a classifier. The experimental results show that the combination of several \glspl{cnn} in an ensemble grants a performance improvement compared with a single \gls{cnn} model.

In~\citet{bochinski2018deep}, two datasets from different biological environments were captured and analyzed. The first dataset was used to analyze the achievable accuracy of the \gls{cnn} and how the Cost-Effective Active Learning (CEAL) can be used to minimize the number of required annotations. The second dataset was used to examine the generalization ability of the \gls{cnn} and if the CEAL method can be used to fine-tune the system to adapt to the characteristics of this new data. 

\citet{plonus2021automatic} suggest using capsule neural networks combined with probability filters to address the dataset shift caused by different plankton imaging instruments. The idea of Capsule neural networks is to form groups of neurons (capsules) that learn the specific properties of the object (e.g. plankton) in the image. The authors argue that the capsule neural networks are less sensitive to the changes in the field conditions and therefore able to adapt to different data distributions.
\citet{guo2022cdfm} proposed a cross‐domain few‐shot learning model for instrument-agnostic plankton recognition. Similarly to transfer learning, the model is first trained on the source domain with a large amount of training data and then adapted to the target domain using fine‐tuning. In addition, graph neural network-based meta-learning is applied to learn a feature distance metric capable of recognizing plankton species in the target dataset with a very limited amount of labeled data.

Domain shifts between the plankton image datasets or imaging instruments have not been widely studied. Most works focus on fine-tuning the recognition models trained on one dataset to new datasets using transfer learning. While the transfer learning reduces the amount of manual labeling needed for new datasets, it does not fully solve the problem of multiple domains. Labeled training data are still needed for all datasets, and the recognition models need to be fine-tuned for each, requiring expertise in machine learning and computing resources. A more general model can be obtained by using ensemble learning with submodels learned on different datasets if training data on each dataset (imaging instrument) is available. More sophisticated approaches to plankton image domain adaptation include the capsule neural networks and meta-learning.

\subsection{Challenge 5: Previously unseen classes and unknown particles}
Automated plankton imaging instruments capture images of unknown particles and the class (plankton species) composition varies between geographical regions and ecosystems. \gls{cnn}-based models are known to struggle in open-set settings where the class composition of training data differs from the data for which the trained model is applied. Typical \gls{cnn}-based classification models tend to classify the images from a new class to one of the known classes often with high confidence, and to include new classes to the models, they need to be retrained. These are major problems for plankton recognition as the plankton species vary between different regions and seasons. Retraining a separate model for each dataset is not feasible. Therefore, there is a need for a recognition model that 1) is able to predict when the image contains a previously unknown plankton species (open-set recognition) and 2) can be generalized to new classes without retraining the whole model.

In the case of plankton recognition, the open-set problem is often formulated as an anomaly detection problem where the model is trained to both correctly classify the known classes and to filter abnormal classes by training the model to produce high and low entropy distributions for the normal classes and abnormal classes respectively. \citet{pastore2020annotation} proposed a semi-automatic method to handle the previously unseen plankton classes by utilizing anomaly detection combined with expert verification. Both one-class SVM and a new neural network-based method called Delta-Enhanced Class (DEC) detector were considered. The DEC detector utilizes absolute differences between the feature vectors of an input image and random images from a known class as additional input to predict whether the input image is from the known class or anomaly. \citet{varma2020autonomous} proposed $L_1$-norm tensor-conformity curation to remove outliers (non-plankton or misclassified images) from the training data. The idea is to measure the conformity of the images using $L_1$-norm subspaces~\citep{tountas2019conformity}. \citet{conradt2022automated} brought up the high intra-class and low inter-class variation of plankton morphology, and spatio-temporal changes in the plankton community as the main causes for the need to frequently validate the results from automatic recognition. The proposed remedy is a dynamic optimization cycle in which the model is updated based on manual-validation results.

\citet{pu2021anomaly} proposed a loss function that contains three loss terms to detect the anomalies and to maintain the classification accuracy for the images belonging to the normal classes by incorporating the expected cross-entropy loss, the expected Kullback-Leibler (KL) divergence, and the Anchor loss. The model was tested on classes of plankton images containing also bubbles or random suspending particles. \citet{walker2021improving} utilized a large background set of images that do not belong to the target classes (classes to be recognized) and hard negative mining to find images that are more likely to cause false negatives. The training set was then complemented with these challenging images to improve the classifier's ability to recognize when the images are from novel classes. While promising results were obtained on open-set plankton recognition the method requires that a labeled background set is available which limits the usability of the method.

Another approach to tackle the open-set problem is to utilize similarity metric learning. The aim of metric learning is to obtain image embedding vectors that model the similarity between images. It is commonly utilized in person \citep{ye2021deep} and animal re-identification \citep{nepovinnykh2020siamese}, as well as content-based image retrieval \citep{dubey2021decade}, but has been also successfully applied to plankton classification \citep{teigen2020leveraging,badreldeen2022open}. 
A simple approach to implement a recognition method is to construct a gallery set of known species and use the learned similarity metric to compare query images to the gallery images. The similarity in this context corresponds to the likelihood that the images belong to the same class. This further allows defining a threshold value for similarity enabling open-set classification: if no similar images are found in the gallery set, the query image is predicted to belong to an unknown class. Furthermore, new classes can be added by simply including them in the gallery set as the model does not necessarily need to learn class-specific image features.

The most common approaches for deep metric learning include triplet-based learning and classification-based metric learning. The first approach learns the metric by sampling image triplets with anchor, positive, and negative examples \citep{hoffer2015deep}. The loss function is defined in such a way that the distances (similarity) from the embeddings of the anchors to the positive samples are minimized, and the distances from the anchors to the negative samples are maximized. The second approach approximates the classes using learned proxies \citep{movshovitz2017no} or class centers \citep{deng2019arcface} that provide the global information needed to learn the metric. This makes it possible to formulate the loss function based on the softmax loss and allows to avoid the challenging triplet mining step.

\citet{teigen2020leveraging} studied the viability of few-shot learners in correctly classifying plankton images. A Siamese network was trained using the triplet loss and used to determine the class of a query image. Two scenarios were tested: the multi-class classification and the novel class detection. A model trained to distinguish between five classes of plankton using five reference images from each class was able to achieve reasonable accuracy. 
In the novel class detection, however, the model was able to filter out only 57 images out of 500 unknowns. 

\citet{badreldeen2022open} utilized the angular margin loss (ArcFace)~\citep{deng2019arcface} instead of triplet loss to address the high cost of the triplet mining step. Furthermore, Generalised Mean pooling (GeM)~\citep{radenovic2018fine} was applied to aggregate the deep activations to rotation and translation invariant representations. ArcFace uses a similarity learning mechanism that allows distance metric learning to be solved in the classification task by introducing the Angular Margin Loss. This allows straightforward training of the model and only adds negligible computational complexity. The metric learning-based method was shown to outperform the model utilizing OpenMax~\citep{bendale2016towards} layer in open-set classification of plankton. One of the main benefits of the method is that it generalizes well to new classes added to the gallery set without retraining. This makes it straightforward to apply the model to new datasets with only partly overlapping plankton species composition.

Plankton species vary in different locations and seasons, thus, it is common that a recognition model should be adapted to or retrained for the new situation at some point. Retraining a separate model for each situation is infeasible, and continual or online training of the model would be challenging for online monitoring applications. Therefore, an effective remedy would be to treat it as an open-set recognition problem, solve it with the modern methods anomaly detection or metric learning, and take care of the model's capability to generalize to new data without the need to retrain the whole model.

\subsection{Challenge 6: Label uncertainty}
The plankton image label uncertainty is caused by the difficulty of manually recognizing the species from low-quality images with limited resolution, human error, and high costs preventing the repetition of the manual annotation by multiple experts. \citet{culverhouse2003experts} identified four main reasons for the incorrect labeling of plankton images: 1) the limited short-term memory of humans, 2) fatigue, 3) recency effects, i.e., labeling is biased towards the most recently seen labels, and 4) positivity bias, i.e., labeling is biased by the expert's expectations to the content of sample. Labels provided by sixteen human experts (marine ecologists and harmful algal bloom monitoring specialists) on microscopy images of dinoflagellates (6 classes) were analyzed. The results showed that only 67 to 83\% self-consistency and 43\% consensus between experts was obtained. Experts who where routinely labeling the selected classes were able to achieve 84 to 95\% labeling accuracy. \citet{culverhouse2007human} brought up several important points related to labeling algae. The presented performance figures do not represent the state-of-the-art of automatic approaches, but improvements would be beneficial for both alternatives. Human expert judgements would benefit from peer review and inter-expert calibration to remove human bias. To improve the automatic solutions, the errors of both man and machine would require further attention. Global reference databases with validated samples and representative coverage of the morphological and physiological characteristics in nature would be beneficial for training and evaluation purposes. In addition, \citet{solow2001estimating} noted that the taxonomic counts of classified individuals are biased when there are errors in classification. A straightforward method for correcting for the bias was proposed based on the classification probabilities of the classifier.

Image filtering has been proposed to address label uncertainty in plankton image data. The idea is to discard images for which the recognition model is uncertain, and therefore, more likely to produce erroneous labels. For example, \citet{faillettaz2016imperfect} utilized a probabilistic \gls{rf} for classification, and obtained class probabilities were used to detect and ignore images for which the classifier is uncertain. \citet{luo2018automated}, \citet{plonus2021automatic}, and \citet{kraft2022towards} utilized similar approach for \gls{cnn}-based recognition models. \citet{luo2018automated} used a separate fully annotated validation set to set class-specific probability thresholds for filtering. \citet{plonus2021automatic} proposed a pipeline for tailoring filtering thresholds to the research question of interest by allowing to select between high precision and high recall. \citet{kraft2022towards} evaluated a \gls{cnn}-based model with class-specific probability thresholds on operational use.

Related to the label uncertainty, quantification methods have been proposed for plankton image data analysis. The basic idea is to estimate the class distribution directly. While mislabeled samples cause noise to the training data for classification methods, the class distributions are often close to correct. 
\citet{sosik2007automated} used a quantification method to estimate the abundance of different taxonomic groups of phytoplankton. Utilizing a combination of image feature types including size, shape, symmetry, and texture characteristics, plus orientation invariant moments, diffraction pattern sampling, and co-occurrence matrix statistics proposed. Statistical analysis was used to estimate category-specific misclassification probabilities for accurate abundance estimates and for quantification of uncertainties in abundance estimates.
\citet{beijbom2015quantification} analyzed several quantification methods on a time-series dataset of plankton samples. These included unsupervised and supervised quantification. In unsupervised quantification, the dataset shift is assumed to be a pure class-distribution shift. Alternatively, the dataset shift is assumed to be ‘small’ and the unlabeled set of target samples is used to align the internal feature representation of a machine learning algorithm. In supervised quantification, no explicit assumptions are made on the dataset shift, but it is assumed that a small amount of samples are available in the target domain. \citet{gonzalez2017validation} proposed a methodology to assess the efficacy of learned models, which takes into account the fact that the data distribution (the plankton composition of the sample) might vary between the training phase and the testing phase. Their approach used validation-by-sample. They proposed using the sample as the basic unit instead of the individuals to predict the abundance of the different plankton groups.
Thus, model assessment processes require groups of samples with sufficient variability to provide precise error estimates. \citet{gonzalez2019automatic} used a transfer learning approach where deep image features as input for the 
quantification algorithm to estimate the distribution of each class in an unknown water sample was proposed. 
\citet{orenstein2020semi} proposed a semi-automatic pipeline where a small subset of images were manually labeled to estimate the dataset shift and use this information to correct the quantification estimate. 

Supervised machine learning and particularly the performance evaluation of a recognition model relies on the correctness of the class labels. However, visual recognition of a number of plankton species from low-quality images is difficult and using expert panels becomes practically infeasible if the aim is to produce large datasets. The proposed remedies include exclusion of images that have high label uncertainty or focusing on the actual quantity of interest if it is not plankton recognition. Alternative ways to solve this challenge would be to focus on few-shot learning with ground truth validated by an expert panel and pay special attention to model generalisability, or to use generative models.

\subsection{Challenge 7: Large image size variation}
\begin{figure}[h] 
     \centering
   \includegraphics[width=0.8\textwidth]{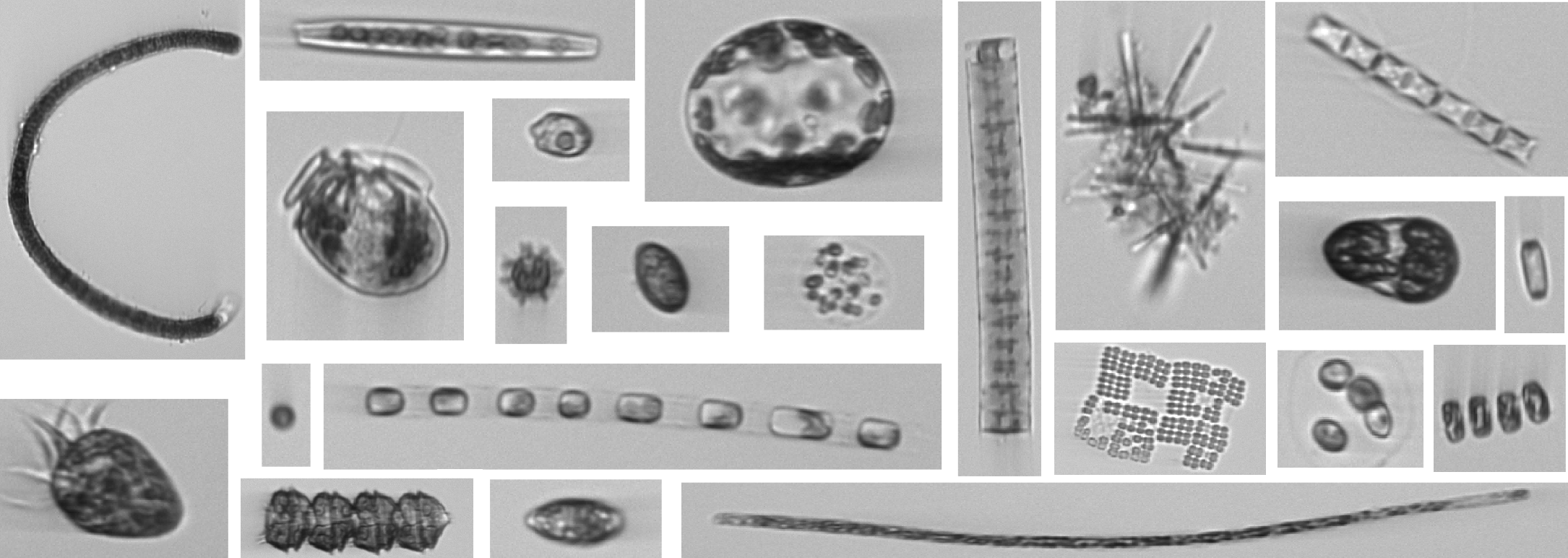}
   \caption{Plankton images with different sizes and aspect ratios.~\citep{burevs2021plankton}.}
   \label{fig:imsize}
 \end{figure}

Most plankton datasets have extreme variation in image size. Fig.~\ref{fig:imsize} shows example images obtained using \gls{ifcb}. Typical CNN-based image classifiers require the input image to have a predefined size. Therefore, image resizing has been used as a pre-processing step for datasets with varying height and width of images \citep[e.g.][]{dai2016hybrid, kuang2015deep}.

On a general level, the resizing can be done in two ways: by forgoing aspect ratio \citep[e.g.][]{al2015performance, sanchez2019diatom} or by maintaining the aspect ratio \citep[e.g.][]{dai2016zooplanktonet,correa2017deep,gonzalez2019automatic}. In the first approach, stretching is needed for images whose aspect ratio does not match the target aspect ratio. This will change the shape of the objects in the image which may affect the feature extraction or learning. In the second approach, images are typically resized based on the length of their longest side and padded with a single color to make the image size correct. \citet{eerola2020towards} evaluated various ways to implement the padding and padding with the mode of the image (the most common color in the image typically corresponding to the background color) produced the best results on \gls{ifcb} data.
Both approaches (forgoing and maintaining aspect ratio) have been utilized in plankton recognition. However, there exists little comparison between them. \citet{dai2016zooplanktonet} tested various resizing methods were tested on zooplankton images and the best accuracy was obtained by maintaining the aspect ratio while scaling. 
On the other hand, \citet{jindalplankton} found little to no difference on performance between the approaches despite images appearing distorted when forgoing aspect ratio.

Various other ways to obtain fixed-size images have been proposed. In the method proposed by \citet{hoflagellates}, a fixed input image size was chosen and the images were either cropped or padded with zeros to adjust them to the correct size. \citet{schroder2018low} proposed to crop the images to their tight bounding box and pad to a square with a minimum edge length of 128 pixels. Images larger than 128 pixels were shrunken to the same size. 
\citet{ellen2019improving} resized images larger than the target size thus losing some detail. Images smaller than the target size were resized by padding and, therefore, the object size remained the same. 
\citet{lumini2019deep, lumini2019ocean, lumini2019deepcoral} compared the two different strategies: 1) resizing all images to a common size and 2) resizing only images that were larger than the input size and using padding for the smaller images. The results showed that the first method produced a better classification result in most of the datasets and models.

All methods to produce fixed-size images from original plankton images with a large size variation result in some degree of information loss or image distortions. Information on the size of the plankton is lost during the resizing, small details disappear if images are heavily downscaled, and only part of the object is seen if cropping is used. \citet{ellen2019improving} partially solved this problem by providing the size information as metadata (additional features) for the classifier while still using resized versions of images as the main input for the \gls{cnn}. 
Metadata is used as an input for the network besides image data, and they are processed independently by separate parts of the network. The outputs of both subnetworks are concatenated together and processed by fully connected layers. Results showed that metadata was useful for classification accuracy.

To truly solve the problem with the varying image size and aspect ratio, the \gls{cnn} architecture needs to be modified so that it can process images with multiple sizes. This can be achieved, e.g. by combining scale-invariant and scale-variant features to devise a multi-scale \gls{cnn} architecture \citep{van2017learning}.
\citet{py2016plankton} proposed an inception module that allows to use multiple scaled versions of the original image with different sizes as the input for \gls{cnn}. By selecting different strides for each scale, the computed feature maps have the same size for all scales and can be concatenated to a single set of multi-scale features. 
The proposed method was shown to outperform the method with a single fixed-size input.

\citet{burevs2021plankton} compared various modifications of the baseline \gls{cnn} on plankton recognition with high variation in image size. These include Spatial Pyramid Pooling (SPP)~\citep{he2015spatial}, using image size as metadata, patch cropping and multi-stream \glspl{cnn}. SPP allows the training of a single CNN with multiple image sizes in order to obtain higher scale invariance by pooling the features produced by the convolutional layer to a fixed-length vector required by the fully connected layers. The metadata was used as described by \citet{ellen2019improving}. 
The patch cropping technique divides images into fixed-size patches that are classified separately. The final recognition is done by averaging the resulting score vectors. Multi-stream \gls{cnn} utilizes a similar approach but uses multiple different networks trained for different image sizes and aspect ratios. The best plankton recognition accuracy was obtained using a multi-stream network combining two models with different input aspect ratios and patch cropping.

Most plankton datasets have significant variation in image sizes and aspect ratios. Common CNN-based image classifiers require that the input images have a constant size. In this case, image resizing is used and it is necessary to consider what to do with the aspect ratio and whether metadata about the image size provides an advantage when complementing the fixed-size images. However, a more general remedy would be to use a multi-scale CNN with an appropriate architecture as the recognition model.

\subsection{Challenge 8: Low or varying image quality}
To improve the classification accuracy on low-quality images various preprocessing steps have been proposed. These include discarding bad quality images~\citep{raitoharju2016data}, image segmentation~\citep{kecceli2017classification}, and denoising~\citep{cheng2019enhanced}.

Low quality images can be discarded in different ways. \citet{raitoharju2016data} manually removed low-quality images from the dataset before training the recognition model. Moreover, the remaining images were cropped to remove artifacts mainly appearing close to image borders. 
\citet{coltelli2014water} filtered out out-of-focus images before the feature extraction. The out-of-focus detection was done by fitting color histograms in a \gls{gmm}. If the distribution contained two components (background and plankton), the image was considered to be in-focus.

Some studies suggest segmenting the images as a preprocessing step to discard non-plankton pixels from the images. For example, \citet{kecceli2017classification} used Otsu's thresholding method~\citep{otsu1975threshold} for segmentation and pixels outside the obtained segmentation map are set to zero.

\citet{cheng2019enhanced} applied texture enhancement together with background suppression before the classification step. Enhanced images were shown to produce a slightly higher recognition accuracy than the images without enhancement.
\citet{ma2021super} proposed to use modern \gls{cnn}-based super-resolution techniques to improve the plankton image quality. The EDRN super-resolution architecture~\citep{lim2017enhanced} was combined with the contextual loss~\citep{mechrez2018contextual}, and was shown to produce high-quality images. However, the effect on plankton recognition accuracy was not assessed.

Many real-world computer vision applications have to deal with low-quality images and plankton recognition is no exception. A wealth of image preprocessing approaches exist and in the case of plankton images, at least exclusion of bad images, denoising and image segmentation have been proposed. A more profound way would be to adopt image reconstruction methods, but from the practical perspective of plankton recognition, the simpler methods can be considered as sufficient and data augmentation is commonly used to introduce additional variation to the data.

\subsection{Challenge 9: Massive amount of data}
Massive data volumes obtained by modern imaging instruments motivate to develop computationally efficient solutions that are able to analyse data in real time. However, the computation time is rarely considered in plankton recognition literature. Most works related to the challenge consider lightweight CNN architectures. For example, shallow TANet~\citep{li2019tanet} was shown to outperform competing methods in computing time without sacrificing accuracy on the Kaggle dataset.

\citet{zimmerman2020embedded} proposed an embedded system for \textit{in situ} deployment of plankton microscope with real-time recognition system. Due to the limited computation resources and computation time limitations, \gls{cnn}-based recognition methods were considered unsuitable and a faster feature-engineering based approach was proposed with reduced recognition accuracy.

The computation time is an especially big issue with holographic imaging that traditionally relies on computationally heavy reconstruction operations to process the raw data. 
To address this end-to-end \gls{cnn} methods for plankton recognition that take the raw holographic data as input have been proposed \citep{guo2021automated,zhang2021automatic}. This way the reconstruction step can be completely avoided. \citet{guo2021automated} and \citet{zhang2021automatic} showed that \glspl{cnn} are able to learn the image features for the plankton recognition from the raw data speeding up the processing significantly.

Online monitoring of plankton with modern imaging equipment produces huge amounts of images. The related image analysis requires either \gls{hpc} resources in the cloud or local (edge) computing with shallow CNN architectures. In most cases, the recognition model training has to be performed in a \gls{hpc} environment after which at least the lightweight models can be deployed for local execution.

\section{Summary and future directions\label{Future directions}} 

In this paper, a comprehensive survey of challenges and existing solutions for automatic plankton recognition was provided. We identified nine challenges that complicate the introduction of automatic plankton recognition methods to operational use: 1) the limited amount of training data for less common species, 2) large class imbalance, 3) fine-grained nature of the recognition task, 4) domain shift between imaging instruments, 5) presence of previously unseen classes and unknown particles, 6) uncertainty in expert labels, 7) large variation in image size, 8) low or varying image quality, and 9) massive data volumes.  
While most of the considered challenges are common in a wide variety of machine learning applications, plankton recognition has its specific characteristics including highly imbalanced image datasets, extreme variation in image size, limitations in image quality, and a shortage of qualified experts to visually annotate the images. 

\begin{figure}[h] 
     \centering
   \includegraphics[width=1\textwidth]{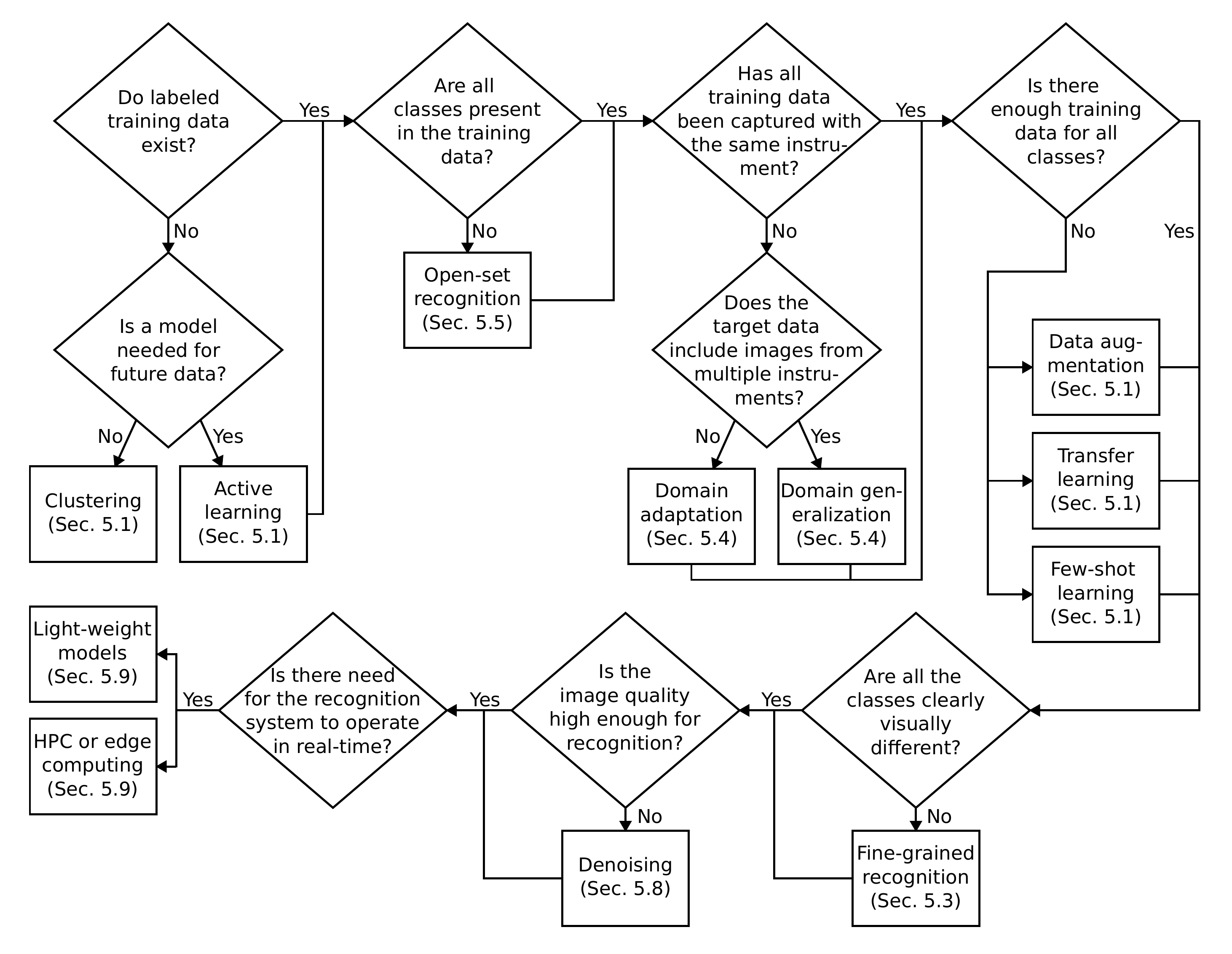}
   \caption{Summary flowchart of challenges and solutions.}
   \label{fig:summary}
\end{figure}

Fig.~\ref{fig:summary} shows a flowchart summarizing the challenges and approaches to solve them. Given a new plankton image dataset, the flowchart provides a simple pipeline to identify the problems related to the dataset as a series of yes-no questions. Furthermore, references to the sections in this paper providing the detailed descriptions are provided to find the existing techniques to tackle the problems and to automate the analysis of the dataset. 

Some of the challenges, especially the limited amount of training data, have been rather extensively studied. While this problem cannot be considered solved, relatively high classification accuracies have been obtained with limited amounts of training images for certain classes. On the other hand, some of the other challenges have not been widely considered in plankton recognition literature. These include the domain shift between different image sets, presence of previously unseen classes and unknown particles, uncertainty in expert labels, and massive data volumes. The reasons for this vary. Most of the research has focused on improving classification accuracy and computation time has not been seen as an issue. Furthermore, the majority of the method development has been done for a fixed set of species and one imaging instrument, thus, there has been no need to address the domain shift or open-set problem.

One notable problem in plankton recognition is the lack of publicly available general-purpose plankton image datasets with an evaluation protocol making it possible to compare different plankton recognition methods in a fair and reliable manner. The vast majority of the research either has focused on private in-house datasets or is based on custom evaluation protocol and dataset splits on publicly available datasets. This makes it impossible to compare the accuracies between different studies making it challenging to select the best practices for future research. This slows down the progress in the plankton recognition method development. Therefore, there is a need for a publicly available plankton dataset with a predetermined evaluation protocol and preferably multiple subsets captured with different imaging instruments to allow quantitative evaluation of the advances in general (device-agnostic) plankton recognition.

Another important problem limiting the wider utilization of automatic plankton recognition is the difficulty of collecting training images to exhaust all the possible classes. It is not realistic to construct a labeled training set consisting of all the plankton species and non-plankton particles that the imaging instrument is capable of capturing in a certain location. Moreover, varying plankton species composition between different geographical regions and ecosystems limits the possibility to apply working recognition models to new locations and datasets. Even a classification model developed and trained for one imaging instrument and one geographic location struggles if new species appear, for example, due to seasonal changes. A more realistic scenario is to aim for open-set plankton recognition models that are able to identify when the images belong to previously unseen classes and either reject them or process them further by, for example, clustering. Open-set recognition is an active research topic in machine learning (see, for example, \citet{geng2020recent}).

Large variation between plankton image datasets with different species compositions and imaging instruments can be considered not only a challenge but also an opportunity. While it is very difficult to develop one general-purpose algorithm for imaging instrument-agnostic plankton recognition, modern domain adaptation methods have the potential to enable the joint utilization of different datasets. This would allow adapting the classification model to new datasets with a reasonable amount of manual work. Domain adaptation has already been successfully applied to various other machine learning applications, such as general object recognition~\citep{wilson2020survey}.
Domain adaptation can be considered a special case of transfer learning that mimics the human vision system and utilizes a model trained in one or more source domains to a different (but related) target domain. Domain adaptation can be utilized to reduce the effect of a large domain shift between different datasets and the lack of labeled training data.

The relatively large pool of different plankton image datasets motivates to further utilize domain generalization and meta-learning to obtain an imaging instrument agnostic recognition model. In meta-learning, multiple datasets and tasks are used to ``learn how to learn'' the recognition model. The idea is to automate the creation of the entire machine learning pipeline end-to-end including the search for the model architecture, hyperparameters, and learning the model weights. Domain generalization refers to learning domain-independent (in this case imaging instrument-independent) feature representations that can be then applied to any dataset. Domain generalization has a wide variety of different applications and it has become an increasingly studied problem in machine learning (see the recent survey in \citet{wang2022generalizing}). Recent progress in such methods has opened novel possibilities to aim towards a universal plankton recognition system that is able to adapt to different environments, with dramatically different plankton populations and varying imaging instruments, promoting the wider utilization of automatic plankton recognition for aquatic research.

\section*{Authors contribution}
\textbf{Tuomas Eerola}: Literature search, Writing - prepare original draft, review \& editing, Visualization, Supervision.
\textbf{Daniel Batrakhanov}: Literature search, Writing - prepare original draft, review \& editing, Visualization.
\textbf{Nastaran Vatankhah Barazandeh}: Writing - prepare original draft, Visualization.
\textbf{Kaisa Kraft}: Writing - review \& editing.
\textbf{Lumi Haraguchi}: Writing - review \& editing.
\textbf{Lasse Lensu}: Writing - review \& editing, Supervision.
\textbf{Sanna Suikkanen}: Writing - review \& editing, Project administration.
\textbf{Jukka Sepp\"al\"a}: Writing - review \& editing.
\textbf{Timo Tamminen}: Writing - review \& editing.
\textbf{Heikki K\"alvi\"ainen}: Writing - review \& editing, Supervision, Project administration.

\section*{Declaration of Competing Interest}
The authors declare that they have no known competing financial interests or personal relationships that could have appeared to influence the work reported in this paper.

\section*{Data availability}
No new data was collected for this survey.

\section*{Acknowledgements}
The research was carried out in the FASTVISION and FASTVISION-plus projects funded by the Academy of Finland (Decision numbers 321980, 321991, 339612, and 339355). Lumi Haraguchi was supported by OBAMA-NEXT (grant agreement no. 101081642), funded by the European Union under the Horizon Europe program.

\clearpage
\appendix
\section{Methods}
\vspace{-20pt}
\begin{table}[b!!!]
    \begin{minipage}{\textwidth}
    \renewcommand*\footnoterule{} 
    \renewcommand*{\thempfootnote}{\arabic{mpfootnote}} 
    \centering
    \caption{Summary of the handcrafted image features used for recognizing plankton.}
    \label{table:feature}
    \resizebox{\textwidth}{!}{%
    \begin{tabular}{llll}
        \toprule
        Category & Features & Publications\\
        \midrule
        \multirow{20}{*}{\makecell[l]{Texture\\ features}} 
            & \multicolumn{1}{l}{\makecell[l]{First order statistical descriptors \\ \citep{julesz1962visual, pratt2007digital}}}
                & \multicolumn{1}{l}{\makecell[l]{
                    \citet{embleton2003automated},
                    \citet{luo2005active},
                    \citet{blaschko2005automatic},\\
                    \citet{kramer2005identifying},
                    \citet{lisin2006image},
                    \citet{sosik2007automated},\\
                    \citet{bell2008assessment},
                    \citet{gorsky2010digital},
                    \citet{kramer2010system} \\
                    \citet{ye2011bayesian},
                    \citet{schulze2013planktovision},
                    \citet{zetsche2014imaging},\\
                    \citet{ellen2015quantifying},
                    \citet{verikas2015integrated},
                    \citet{faillettaz2016imperfect},\\
                    \citet{corgnati2016looking},
                    \citet{hirata2016plankton},
                    \citet{lai2016high},\\
                    \citet{zheng2017automatic},
                    \citet{bueno2017automated},
                    \citet{solano2018radiolarian},\\
                    \citet{wacquet2018combination},
                    \citet{shan2020automated},
                    \citet{arje2020human},\\
                    \citet{guo2021real},
                    \citet{mirasbekov2021semi},
                    \citet{liu2021plankton}}} \\
                    \cline{2-3}  \addlinespace
                & \multicolumn{1}{l}{\makecell[l]{Second order statistical descriptors\\ (Haralick features \& Co-occurrence \\ matrix (COM)) \citep{haralick1973textural}}}
                & \multicolumn{1}{l}{\makecell[l]{
                    \citet{thiel1995automated},
                    \citet{walker2002fluorescence},
                    \citet{hu2005automatic},\\
                    \citet{blaschko2005automatic},
                    \citet{lisin2006image},
                    \citet{hu2006accurate},\\
                    \citet{sosik2007automated},
                    \citet{gorsky2010digital},
                    \citet{schulze2013planktovision},\\
                    \citet{verikas2015integrated},
                    \citet{faillettaz2016imperfect},
                    \citet{corgnati2016looking},\\
                    \citet{hirata2016plankton},
                    \citet{lai2016high},
                    \citet{solano2018radiolarian},\\
                    \citet{bueno2017automated},
                    \citet{campbell2020prince},
                    \citet{shan2020automated},\\
                    \citet{guo2021real},
                    \citet{liu2021plankton},
                    \citet{wei2022microalgae}}} \\
                    \cline{2-3} \addlinespace
            & \multicolumn{1}{l}{\makecell[l]{Binary Gradient Contours\\ (BGC) \citep{fernandez2011image}}}
                & \multicolumn{1}{l}{
                    \citet{zheng2017automatic}} \\
                    \cline{2-3}  \addlinespace
            & \multicolumn{1}{l}{\makecell[l]{Gabor Descriptors\\ \citep{fischer2007self}}}
                & \multicolumn{1}{l}{\makecell[l]{
                    \citet{ellis1997committees},
                    \citet{verikas2015integrated},
                    \citet{bueno2017automated},\\
                    \citet{zheng2017automatic},
                    \citet{sanchez2019diatom},
                    \citet{liu2021plankton},\\
                    \citet{rivas2021fully}}} \\ 
                    \cline{2-3} \addlinespace
            & \multicolumn{1}{l}{\makecell[l]{Local Binary Pattern (LBP)\\ \citep{ojala2002multiresolution}}}
                & \multicolumn{1}{l}{\makecell[l]{
                    \citet{blaschko2005automatic},
                    \citet{lisin2005combining},
                    \citet{lisin2006image},\\
                    \citet{schulze2013planktovision},
                    \citet{chang2016phytoplankton},
                    \citet{bueno2017automated},\\
                    \citet{zheng2017automatic},
                    \citet{liu2021plankton}}} \\
                    \cline{2-3} \addlinespace
            & \multicolumn{1}{l}{\makecell[l]{Variogram Function (VF)\\ \citep{zheng2017automatic}}}
                & \multicolumn{1}{l}{\makecell[l]{
                    \citet{zheng2017automatic}}} \\
                    \cline{2-3}  \addlinespace
            & \multicolumn{1}{l}{\makecell[l]{Fourier Descriptors on texture\\\citep{cosgriff1960idntification}}}
                & \multicolumn{1}{l}{\makecell[l]{
                    \citet{ellis1997committees},
                    \citet{walker2002fluorescence},
                    \citet{kramer2005identifying},\\
                    \citet{mosleh2012preliminary}}} \\
        \midrule

        \multirow{15}{*}{\makecell[l]{Shape\\ features}}
             & \multicolumn{1}{l}{\makecell[l]{Fourier Descriptors on contours\\ \citep{cosgriff1960idntification}}} 
                & \multicolumn{1}{l}{\makecell[l]{
                    \citet{thiel1995automated},
                    \citet{ellis1997committees},
                    \citet{tang1998automatic},\\
                    \citet{embleton2003automated},
                    \citet{davis2004real},
                    \citet{blaschko2005automatic},\\
                    \citet{kramer2005identifying},
                    \citet{luo2005active},
                    \citet{zhao2005binary},\\
                    \citet{hu2006accurate},
                    \citet{tang2006binary},
                    \citet{rodenacker2006automatic},\\
                    \citet{zhao2009bagging},
                    \citet{zhao2010binary},
                    \citet{kramer2010system},\\
                    \citet{luo2011automatic},
                    \citet{dimitrovski2012hierarchical},
                    \citet{li2013pairwise},\\
                    \citet{verikas2015integrated},
                    \citet{lauffer2017morphological},
                    \citet{liu2017efficient},\\
                    \citet{liu2020shape}}}\\
                    \cline{2-3} \addlinespace
            & \multicolumn{1}{l}{\makecell[l]{General Geometric features\\ and derived descriptors}} 
                & \multicolumn{1}{l}{\makecell[l]{
                    \citet{gorsky1989autonomous},
                    \citet{thiel1995automated},
                    \citet{ellis1997committees},\\
                    \citet{embleton2003automated},
                    \citet{walker2002fluorescence},
                    \citet{luo2003learning},\\
                    \citet{luo2004recognizing},
                    \citet{davis2004real},
                    \citet{lisin2005combining},\\
                    \citet{luo2005scaling},
                    \citet{blaschko2005automatic},
                    \citet{kramer2005identifying},\\
                    \citet{luo2005active},
                    \citet{lisin2006image},
                    \citet{hu2006accurate},\\
                    \citet{tang2006binary},
                    \citet{sosik2007automated},
                    \citet{rodenacker2006automatic},\\
                    \citet{zhao2009bagging},
                    \citet{zhao2010binary},
                    \citet{gorsky2010digital},\\
                    \citet{kramer2010system}
                    \citet{ye2011bayesian},
                    \citet{mosleh2012preliminary}
                    \citet{li2013pairwise},\\
                    \citet{drewsjr2013microalgae},
                    \citet{schulze2013planktovision},
                    \citet{zetsche2014imaging},\\
                    \citet{bi2015semi},
                    \citet{ellen2015quantifying},
                    \citet{verikas2015integrated},\\
                    \citet{faillettaz2016imperfect},
                    \citet{corgnati2016looking},
                    \citet{hirata2016plankton},\\
                    \citet{lai2016high}, 
                    \citet{dyomin2017fast},
                    \citet{zheng2017automatic},\\
                    \citet{bueno2017automated}, 
                    \citet{solano2018radiolarian},
                    \citet{wacquet2018combination},\\
                    \citet{shan2020automated},
                    \citet{liu2020shape},
                    \citet{arje2020human},\\
                    \citet{guo2021real},
                    \citet{mirasbekov2021semi},
                    \citet{liu2021plankton}}} \\
        \bottomrule
    \end{tabular}}
    \end{minipage}
\end{table}

\begin{table}[h]
    \begin{minipage}{\textwidth}
    \renewcommand*\footnoterule{} 
    \renewcommand*{\thempfootnote}{\arabic{mpfootnote}} 
    \centering
    \caption{Summary of the handcrafted image features (continue).}
    \label{table:feature2}
    \resizebox{\textwidth}{!}{%
    \begin{tabular}{llll}
        \toprule
        Category & Features & Publications\\
        \midrule
        \multirow{20}{*}{\makecell[l]{Shape\\ features}} 
            & \multicolumn{1}{l}{\makecell[l]{Symmetry measures (Hausdorff\\ distance) \citep{du2002automatic}}}
                & \multicolumn{1}{l}{\makecell[l]{
                    \citet{lisin2005combining},
                    \citet{sosik2007automated},
                    \citet{guo2021real}}}\\
                    \cline{2-3} \addlinespace
            & \multicolumn{1}{l}{\makecell[l]{Freeman contour code features\\ \citep{freeman1961encoding}}}
                & \multicolumn{1}{l}{\makecell[l]{
                    \citet{rodenacker2006automatic}}}\\
                    \cline{2-3} \addlinespace
            & \multicolumn{1}{l}{\makecell[l]{Granulometries \citep{kingman1975g}}}
                & \multicolumn{1}{l}{\makecell[l]{
                    \citet{tang1998automatic},
                    \citet{luo2003learning},
                    \citet{luo2004recognizing},\\
                    \citet{davis2004real},
                    \citet{kramer2005identifying},
                    \citet{luo2005scaling},\\
                    \citet{luo2005active},
                    \citet{zhao2005binary},
                    \citet{lisin2006image},\\
                    \citet{hu2006accurate},
                    \citet{tang2006binary},
                    \citet{zhao2009bagging},\\
                    \citet{zhao2010binary},
                    \citet{li2013pairwise},
                    \citet{zheng2017automatic}}}\\
                    \cline{2-3} \addlinespace
            & \multicolumn{1}{l}{\makecell[l]{Diffraction patterns\\ \citep{lendaris1970diffraction}}}
                & \multicolumn{1}{l}{\makecell[l]{\citet{sosik2007automated},
                    \citet{guo2021real}}} \\
                    \cline{2-3} \addlinespace
            & \multicolumn{1}{l}{\makecell[l]{Circular Projection\\ \citep{tang2006binary}}}
                & \multicolumn{1}{l}{\makecell[l]{
                    \citet{zhao2005binary},
                    \citet{tang2006binary},
                    \citet{zhao2009bagging},\\
                    \citet{zhao2010binary},
                    \citet{li2013pairwise}}}\\
                    \cline{2-3} \addlinespace
            & \multicolumn{1}{l}{\makecell[l]{Object density \citep{tang2006binary}}}
                & \multicolumn{1}{l}{\makecell[l]{
                    \citet{zhao2005binary},
                    \citet{tang2006binary},
                    \citet{zhao2009bagging},\\
                    \citet{zhao2010binary},
                    \citet{li2013pairwise}}}\\
                    \cline{2-3} \addlinespace
            & \multicolumn{1}{l}{\makecell[l]{Boundary smoothness\\ \citep{tang2006binary}}}
                & \multicolumn{1}{l}{\makecell[l]{
                    \citet{tang2006binary},
                    \citet{zhao2005binary},
                    \citet{zhao2009bagging},\\
                    \citet{zhao2010binary},
                    \citet{li2013pairwise},
                    \citet{liu2020shape}}} \\
                    \cline{2-3} \addlinespace
            & \multicolumn{1}{l}{\makecell[l]{Zernike Moments (ZM)\\ \citep{khotanzad1990invariant}}}
                & \multicolumn{1}{l}{\makecell[l]{
                    \citet{blaschko2005automatic},
                    \citet{liu2020shape}}}\\
                    \cline{2-3} \addlinespace
            & \multicolumn{1}{l}{\makecell[l]{Affine Curvature Descriptors (ACD)\\ \citep{liu2020shape}}}
                & \multicolumn{1}{l}{\makecell[l]{
                    \citet{liu2020shape}}}\\
                    \cline{2-3} \addlinespace
            & \multicolumn{1}{l}{\makecell[l]{Elliptical Fourier Descriptors (EFD)\\ \citep{kuhl1982elliptic}}}
                & \multicolumn{1}{l}{\makecell[l]{
                    \citet{schulze2013planktovision},
                    \citet{beszteri2018quantitative},
                    \citet{sanchez2019diatom2}}}\\
                    \cline{2-3} \addlinespace
            & \multicolumn{1}{l}{\makecell[l]{Moment Invariants (Hu Moments)\\ \citep{hu1962visual, reiss1991revised}}}
                & \multicolumn{1}{l}{\makecell[l]{
                    \citet{thiel1995automated},
                    \citet{tang1998automatic},
                    \citet{luo2003learning},\\
                    \citet{luo2004recognizing},
                    \citet{davis2004real},
                    \citet{blaschko2005automatic},\\
                    \citet{kramer2005identifying},
                    \citet{luo2005scaling},
                    \citet{luo2005active},\\
                    \citet{zhao2005binary},
                    \citet{lisin2006image},
                    \citet{hu2006accurate},\\
                    \citet{tang2006binary},
                    \citet{sosik2007automated},
                    \citet{rodenacker2006automatic},\\
                    \citet{zhao2009bagging},
                    \citet{zhao2010binary},
                    \citet{kramer2010system},\\
                    \citet{schulze2013planktovision},
                    \citet{hirata2016plankton},
                    \citet{lai2016high},\\
                    \citet{bueno2017automated},
                    \citet{li2013pairwise},
                    \citet{shan2020automated},\\
                    \citet{guo2021real},
                    \citet{liu2021plankton}}} \\
                    \cline{2-3} \addlinespace
            & \multicolumn{1}{l}{\makecell[l]{SHERPA (Shape Recognition,\\ Processing and Analysis)\\ \citep{kloster2014sherpa}}}
                & \multicolumn{1}{l}{\makecell[l]{
                    \citet{beszteri2018quantitative},
                    \citet{liu2021plankton}}} \\
                    \cline{2-3} \addlinespace
            & \multicolumn{1}{l}{\makecell[l]{Hough descriptors\\ \citep{duda1972use}}}
                & \multicolumn{1}{l}{\makecell[l]{
                    \citet{blaschko2005automatic}}} \\
                    \cline{2-3} \addlinespace
            & \multicolumn{1}{l}{\makecell[l]{Shape index \citep{ravela2003multi}}}
                & \multicolumn{1}{l}{\makecell[l]{
                    \citet{lisin2005combining}}} \\                                                
        \midrule
        \multirow{8}{*}{\makecell[l]{Local\\ features}} 
            & \multicolumn{1}{l}{\makecell[l]{Histogram of Oriented Gradients\\ (HOG) \citep{dalal2005histograms}}}
                & \multicolumn{1}{l}{\makecell[l]{
                    \citet{schulze2013planktovision},
                    \citet{bi2015semi},
                    \citet{zheng2017automatic},\\
                    \citet{guo2021real}}}  \\
                    \cline{2-3} \addlinespace
            & \multicolumn{1}{l}{\makecell[l]{Inner-Distance shape context\\ (IDSC) \citep{ling2007shape}}}
                & \multicolumn{1}{l}{\makecell[l]{
                    \citet{zheng2017automatic}}}\\
                    \cline{2-3} \addlinespace
            & \multicolumn{1}{l}{\makecell[l]{Phase congruency descriptors\\ (PCD) \citep{kovesi2000phase, kovesi2003phase}}}
                & \multicolumn{1}{l}{\makecell[l]{
                    \citet{verikas2012phase},
                    \citet{verikas2015integrated},
                    \citet{sanchez2019diatom}}}\\
                    \cline{2-3} \addlinespace
            & \multicolumn{1}{l}{\makecell[l]{Scale Invariant Feature\\ Transform (SIFT)\\ \citep{lowe1999object,lowe2004distinctive}}}
                & \multicolumn{1}{l}{\makecell[l]{
                    \citet{lisin2005combining},
                    \citet{lisin2006image},
                    \citet{tsechpenakis2007image},\\
                    \citet{dimitrovski2012hierarchical},
                    \citet{zheng2017automatic}}}\\
                    \cline{2-3} \addlinespace
            & \multicolumn{1}{l}{\makecell[l]{Speeded Up Robust Features\\ (SURF) \citep{bay2006surf}}}
                & \multicolumn{1}{l}{\makecell[l]{
                    \citet{chang2016phytoplankton}}}\\
        \bottomrule
    \end{tabular}}
    \end{minipage}
\end{table}

\begin{table}[h]
    \centering
    \caption{Summary of the CNN architectures used in the literature. The architectures developed specifically for plankton recognition are shown in bold. It should be noted that many architectures have various versions with different depths (e.g. VGGNet and ResNet). The number of layers and parameters for each architecture are based on the original publication.}
    \begin{tiny}
    \begin{tabular}{lrrp{5.8cm}}
        \hline
        Architecture & \makecell[l]{Number of\\layers} & \makecell[l]{Number of\\parameters}  & Publication \\
        \hline
        \makecell[l]{AlexNet\\ \citep{krizhevsky2012imagenet}} & 8 & 60 M & \citet{py2016plankton,raitoharju2016data,dai2016hybrid,orenstein2017transfer,kecceli2017classification,pedraza2017automated,pedraza2018lights,rodrigues2018evaluation,wang2018transferred,bochinski2018deep,cui2018texture,liu2018teaching,liu2018deep,libreros2018automated,sanchez2019diatom,pardeshi2019classification,cheng2019enhanced,lumini2019deep,du2020plankton,song2020classification,arje2020human,khan2022multiclass,vallez2022diffeomorphic}
    \\
        \makecell[l]{\textbf{ClassyFireNet}}\\ \citep{jindalplankton} & 8 & - & \citet{jindalplankton},
    \\
        \makecell[l]{DenseNet\\ \citep{huang2017densely}} & 121-169 & 7.2-12.8 M & \citet{sanchez2019diatom,lumini2019deep,kloster2020deep,kyathanahally2021deep,vallez2022diffeomorphic}
    \\
        \makecell[l]{GoogleNet\\ \citep{szegedy2015going}} & 22 & 6.9 M  & \citet{jindalplankton,dai2016hybrid,liu2018teaching,liu2018deep,libreros2018automated,cheng2019enhanced,lumini2019deep,khan2022multiclass}
    \\
        \makecell[l]{InceptionV3\\ \citep{szegedy2016rethinking}} & 48 & 23.6 M & \citet{sanchez2019diatom,lumini2019deep,du2020plankton,kloster2020deep,macneil2021plankton,kyathanahally2021deep,vallez2022diffeomorphic}
    \\
        \makecell[l]{MobileNetV2\\ \citep{sandler2018mobilenetv2}} & 53 & 3.4 M & \citet{lumini2019deepcoral,kloster2020deep,kyathanahally2021deep}
    \\
        \makecell[l]{NasNet\\ \citep{zoph2018learning}} & & & \citet{lumini2019deepcoral}
    \\
        \makecell[l]{PyramidNet\\ \citep{han2017deep}} & 110 & 1.7 M & \citet{liu2018teaching,liu2018deep}
    \\
        \makecell[l]{ResNet\\ \citep{he2016deep}} & 18-152 & 11-58 M & \citet{li2016deep,yan2017more,schroder2018low,liu2018teaching,liu2018deep,dunker2018combining,libreros2018automated,sanchez2019diatom,cheng2019enhanced,gonzalez2019automatic,mitra2019automated,lumini2019deep,sun2020few,du2020plankton,teigen2020leveraging,schmarje2021fuzzy,guo2021classification,macneil2021plankton,walker2021improving,kyathanahally2021deep,guo2021classification,pu2021anomaly,xu2022accurate,khan2022multiclass,vallez2022diffeomorphic}
    \\
        \makecell[l]{InceptionResNetV2\\ \citep{szegedy2017inception}} & 164 & 55 M & \citet{kloster2020deep}
    \\
        \makecell[l]{SqueezeNet\\ \citep{iandola2016squeezenet}} & 18 & 1.25 M & \citet{sanchez2019diatom,vallez2022diffeomorphic}
    \\
        \makecell[l]{\textbf{TANet}\\ \citep{li2019tanet}} & 8 & & \citet{li2019tanet}
    \\
        \makecell[l]{VGGNet\\ \citep{simonyan2014very}} & 11-19 & 15.1-25.9 M & \citet{kuang2015deep,li2016deep,yan2017more,hoflagellates,wang2018transferred,liu2018teaching,liu2018deep,sanchez2019diatom,ellen2019improving,cheng2019enhanced,mitra2019automated,lumini2019deep,teigen2020leveraging,du2020plankton,kloster2020deep,varma2020autonomous,macneil2021plankton,qiao2021classification,pu2021anomaly,khan2022multiclass,rachman2022application,conradt2022automated,vallez2022diffeomorphic}
    \\
       \makecell[l]{Xception\\ \citep{chollet2017xception}} & 36 & 22.8 M &  \citet{hoflagellates,kloster2020deep,macneil2021plankton}
    \\
       \makecell[l]{\textbf{ZooplanktoNet}\\ \citep{dai2016zooplanktonet}} & 11 &  & \citet{dai2016zooplanktonet,geraldes2019situ}
    \\
        \makecell[l]{EfficientNet\\ \citep{tan2019efficientnet}} & 49 & 5.3 M & \citet{venkataramanan2021tackling,kyathanahally2021deep}
    \\
        \makecell[l]{SeNet\\ \citep{hu2018squeeze}} & 52 & 25.6 M & \citet{xu2022accurate}
    \\
    \hline
    \end{tabular}
    \end{tiny}
    \label{table:architecture}
\end{table}

\clearpage
\bibliographystyle{elsarticle-harv} 
\bibliography{references}





\end{document}